\documentclass{article}

\usepackage{microtype}
\usepackage{subcaption}
\usepackage{booktabs} % for professional tables

\usepackage{hyperref}

\usepackage[preprint]{icml2026}

\usepackage{amsmath}
\usepackage{amssymb}
\usepackage{mathtools}
\usepackage{amsthm}
\usepackage{subcaption}
\usepackage{wrapfig}
\usepackage{natbib}
\usepackage{graphicx}
\usepackage{bm}
\usepackage{xcolor}  
\usepackage[dvipsnames]{xcolor}
\usepackage[table]{xcolor}
\usepackage[capitalize,noabbrev]{cleveref}

\theoremstyle{plain}

\theoremstyle{definition}

\theoremstyle{remark}

\usepackage[textsize=tiny]{todonotes}

\newcommand{\ModelName}{MarCos}

\icmltitlerunning{Deep Thinking by Markov Chain of Continuous Thoughts}

\begin{document}

\twocolumn[
  \icmltitle{Deep Thinking by Markov Chain of Continuous Thoughts}



  \icmlsetsymbol{equal}{*}
  \icmlsetsymbol{intern}{$\dag$}
  
  \begin{icmlauthorlist}
    \icmlauthor{Jiayu Liu}{yyy,intern}
    \icmlauthor{Zhenya Huang}{yyy}
    \icmlauthor{Xuan Yang}{sch,sch1}
    \icmlauthor{Tianyun Ji}{yyy}
    \icmlauthor{Anya Sims}{comp}
     \icmlauthor{Hao Xu}{xxx}
    \icmlauthor{Enhong Chen}{yyy}
    \icmlauthor{Yee Whye Teh}{comp}
    \icmlauthor{Ning Miao}{sch,sch1}
  \end{icmlauthorlist}

  \icmlaffiliation{yyy}{State Key Laboratory of Cognitive Intelligence, University of Science and Technology of China}
  \icmlaffiliation{comp}{Department of Statistics, University of Oxford}
  \icmlaffiliation{xxx}{Li Auto Inc}
  \icmlaffiliation{sch}{Department of Data Science, City University of Hong Kong}
  \icmlaffiliation{sch1}{Hong Kong Institute of AI for Science, City University of Hong Kong}

  \icmlcorrespondingauthor{Ning Miao}{ningmiao@cityu.edu.hk}
  \icmlcorrespondingauthor{Zhenya Huang}{huangzhy@ustc.edu.cn}

  \icmlkeywords{Machine Learning, ICML}

  \vskip 0.3in
]



\printAffiliationsAndNotice{$^\dag$Work done during Jiayu's visit to the University of Oxford.} % no special notice (required even if empty)

\begin{abstract}
Transformer-based models can perform complicated reasoning by generating reasoning paths token by token.
While effective, this approach often requires generating thousands of tokens to solve a single problem, which can be slow and computationally expensive.
More importantly, it involves a discrete sampling operation at the end of each time step, creating an information bottleneck across time steps. In this work, we propose {\ModelName}, an improvement of the transformer structure that allows fully continuous reasoning at the thought level.
Unlike traditional transformer layers, which focus on refining token predictions at each time step, layers in {\ModelName} map a continuous representation of a stepwise thought to the distribution of the next thought. 
This enables us to achieve multi-step reasoning in a single pass of {\ModelName}.
Preliminary experimental results on synthetic and real-world math tasks show the great potential of {\ModelName}. Notably, we observe that the increased information bandwidth of {\ModelName} elicits the ability of parallel thinking, in contrast to single-threaded thinking in traditional transformers. 
Meanwhile, in real-world math tasks, {\ModelName} achieves more than $10\times$ speedup in wall-clock time with the same level of accuracy. Our code is available at~\url{https://github.com/Ljyustc/MarCos} .
\end{abstract}

\section{Introduction}
Transformer-based models exhibit strong reasoning capabilities by autoregressively generating reasoning paths token by token~\citep{wei2022chain}.
Despite its simplicity and applicability, this token-wise autoregressive generation can lead to inefficient inference and suboptimal reasoning behavior.
To solve a problem, we often need to serially invoke the transformer thousands of times, resulting in significant latency.
Meanwhile, transformers are forced to think and speak simultaneously.
In other words, they need to generate one token immediately after a single forward computation, leaving little room for thoughtful planning~\citep{valmeekam2023planning,kambhampati2024position}.
Moreover, the discrete operation of token selection at the end of each time step heavily restricts the information bandwidth between time steps.
For example, the information generated at the last transformer layer can only be passed to the next step through the sampled tokens, leading to severe information loss.

To mitigate these issues, previous works~\citep{deng2023implicit, deng2024explicit} have attempted to replace discrete tokens in transformers with continuous ones. 
For example, Coconut~\citep{hao2024training} gradually replaces blocks of discrete tokens with fewer continuous tokens. CoLaR~\citep{tan2025think} further introduces an additional loss to align the continuous tokens with the representation of discrete tokens in ground-truth reasoning paths.
Although these methods reduce the number of time steps to reach a final answer, they mainly work by compressing multiple discrete tokens into a continuous one, which means that their latent thinking essentially remains a form of token-based decision-making.
Moreover, most of them adopt a deterministic transition function between continuous tokens during reasoning, which is unsuitable for learning the diverse and multi-modal solution space.
%As a result, none of them achieve the same level of performance compared with explicit CoT.
%\textcolor{red}{Ning: Is it true?}
%However, current reasoning paradigm is fundamentally constrained in two ways. First, within the transformer architecture on which these models are based, each token is produced through a process that entangles internal reasoning (i.e., within intermediate layers) with linguistic realization (i.e., output layer). In other words, the model is asked to ``think while speaking''. This tight coupling forces the model to respond a token immediately after thinking, leaving little room for thoughtful planning beforehand~\citep{levelt1993speaking,dell1992stages}. Second, existing reasoning methods (e.g., test-time scaling up) are fully grounded in explicit linguistic tokens, which forces all reasoning to be conducted in the language space~\citep{hao2024training,tan2025think}. This is contrast with human cognition, where reasoning often relies on implicit and abstract representations in the brain that are only verbalized when necessary for communication~\citep{fedorenko2011functional,fedorenko2024language}.

\begin{figure*}[t]
\centering
\includegraphics[width=0.83\linewidth]{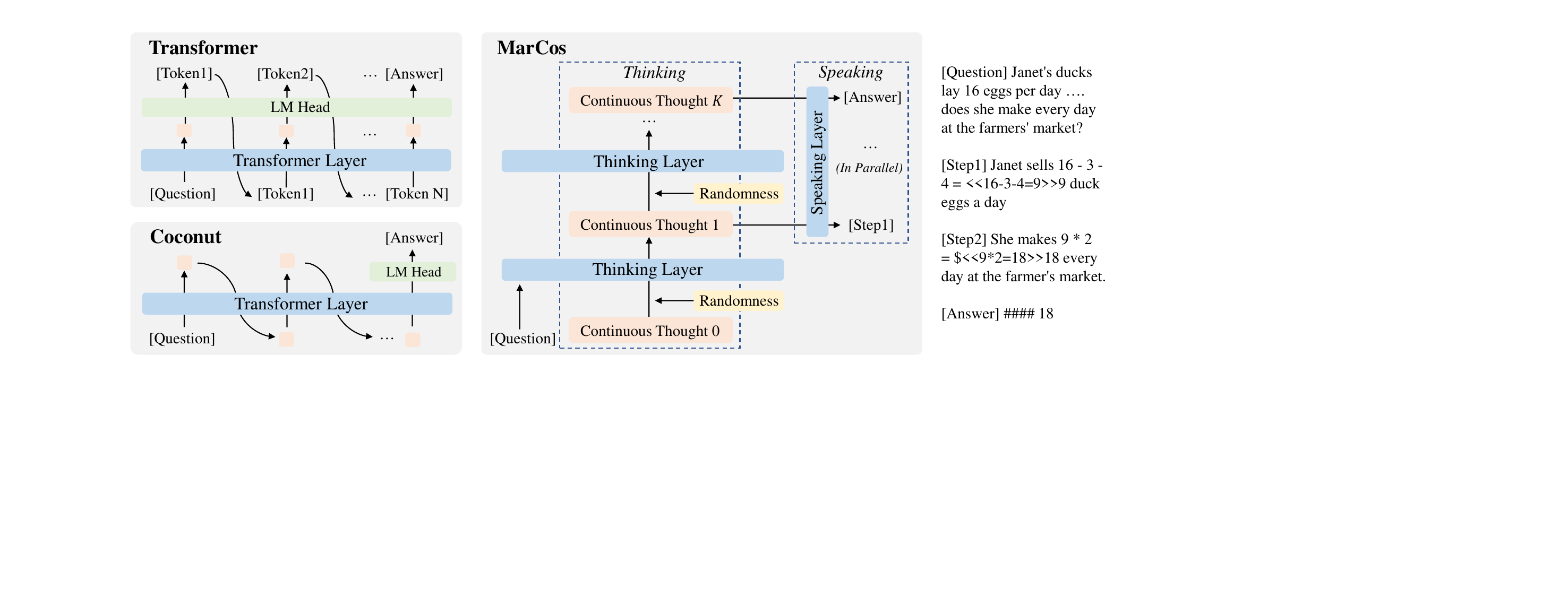}
\caption{Comparison of a traditional transformer, the representative continuous reasoning method Coconut, and our {\ModelName}. In {\ModelName}, the \emph{thinking} process is modeled as iterative transitions of continuous thoughts ($0$ to $K$) with incorporated randomness. The \emph{speaking} process (optionally) translates these thoughts into natural language (e.g., [Step1] to [Answer]).
%\textcolor{red}{We should probably change our model figure into a HMC style figure like the first figure in https://dlab.berkeley.edu/news/brief-primer-hidden-markov-models}
}
%\textcolor{red}{I don't think we should call the thinking module as LLM the figure, because it's a markov process on mind states, instead of a LM. Should maybe call it thinking module or `Markovian thinking module' if we want to emphasize the randomness modelling? Also should we call it \ModelName or cosmas? Or simple CMMS?}
\label{figure_demo}
\vspace{-8pt}
\end{figure*}
In this paper, we propose \textbf{\ModelName}, a variant of transformer that natively supports continuous reasoning at the thought level.
In {\ModelName}, we model the step-by-step reasoning process as a Markov chain of continuous thoughts.
%As shown in Figure~\ref{figure_demo}, starting from an initial thought, {\ModelName} predicts the distribution of thoughts at next step.
Different from traditional transformers, whose layers are in charge of refining the prediction of the next token~\citep{lioubashevski2024looking}, {\ModelName} employs dedicated thinking layers to directly predict the distribution of thoughts at the next reasoning step based on the current thought, as illustrated in Figure~\ref{figure_demo}.
By representing each thought as the states of a group of neurons, {\ModelName} essentially thinks internally by updating neuron states.
To peek into and add supervision on intermediate thoughts, we use a separate speaking layer to translate neuron states to natural language.
This design essentially disentangles the internal \emph{thinking} process from the \emph{speaking} process, which allows the model to fully deliberate before speaking. It also aligns with recent neuroscience findings that language is primarily a tool for communication rather than thought~\citep{fedorenko2024language,fedorenko2016language}.
To model the diverse distribution of possible reasoning paths, we further incorporate a variational module that explicitly controls the randomness of reasoning at the step level. 
With all these designs, we enable {\ModelName} to think completely in a continuous space, which creates the possibility for parallel and more efficient reasoning.

We first evaluate {\ModelName} on synthetic tasks, where we uncover that its superior bandwidth enables parallel thinking, a capability that is difficult to achieve in traditional transformers. Then, on three grade-school level mathematical benchmarks (GSM8K~\citep{cobbe2021training}, SVAMP~\citep{patel2021nlp}, and MultiArith~\citep{roy2015solving}), {\ModelName} \textbf{obtains performance on par with or even superior to transformer-based CoT reasoning}, with a $4.7$\% accuracy improvement on GSM8K and up to $15.7\times$ inference-time speedup. 
This highlights the great potential of our architecture. Beyond accuracy, our analysis reveals the functional independence between contextual processing and randomness determination in variational modules, and uncovers that different randomness dimensions control different speaking properties such as sentence length and step depth. This opens up the possibility of manually controlling reasoning behaviors. 
%\textcolor{red}{Do we have cases for controlling high-level reasoning? May use a generated example, rather than examples with ground truth steps?}
Furthermore, we show that our model remains competitive when applying non-autoregressive (NAR) decoding in the speaking stage, indicating that thinking and speaking play well-separated roles in our architecture. 
In summary, our contributions in this paper are:

\begin{itemize}
    \item We propose {\ModelName}, a new architecture that models reasoning as a Markov chain of continuous thoughts and operates in decoupled \emph{thinking-speaking} stages. 
    %The thinking stage conducts \emph{step-by-step} implicit reasoning in a latent space, while the speaking stage merely translates the internal thoughts into natural language, which can be realized using either autoregressive or non-autoregressive decoding.
    \item We introduce a principled way to model the randomness in reasoning, which yields two key advances: (1) enabling explicit control over the reasoning process, and (2) verifying the feasibility of pre-determining step-level randomness before token generation.
    \item Experiments on synthetic and math tasks show the strong potential of {\ModelName}, which can achieve up to $15.7\times$ faster inference without loss in accuracy.
    %achieves state-of-the-art reasoning performance, improving over the best continuous reasoning model by 8.66\% accuracy and even surpassing token-based CoT by $4.7$\% on GSM8K with up to $15.7\times$ speedup.
\end{itemize}
% We identify and formalize the thinking-speaking entanglement problem in autoregressive LLMs.

% We propose a novel model that introduces an explicit latent thinking phase before verbal output.

% We demonstrate that this decoupled inference paradigm leads to better reasoning performance on complex benchmarks.
\section{Related Work}
% \textcolor{red}{Ning: Need to explain why the previous methods didn't solve the problem, and why our method is needed.}
% \textcolor{red}{Ning: I think we should start directly with continuous / hierarchical reasoning in this part. We already the relationship with explicit CoT in the introduction part.}
% \textcolor{red}{Ning: We may want to include more loosely related related works. Do not need to introduce them separately, but put the same type of them into a single sentence.}
%While chain-of-thought (CoT) prompting~\citep{wei2022chain} has become a dominant paradigm for enhancing reasoning in LLMs, it relies heavily on explicit natural language rationales. 
%Recent work highlights several drawbacks of this paradigm, including verbosity and increased inference cost~\citep{liu2025answer}, potential misalignment between textual rationales and the model’s actual reasoning process~\citep{turpin2023language}, and the divergence from humans' cognitive mechanism~\citep{liumind}. To address these issues, emerging research explores latent reasoning, where reasoning is performed internally in the hidden space rather than being verbalized step by step. 
Various continuous reasoning methods have emerged to mitigate the issues of token-by-token reasoning, which can be categorized into time-axis and depth-axis approaches.
%, including slow inference speed and restricted information bandwidth

\textbf{Time-axis latent reasoning.}
Most time-axis approaches aim to improve inference efficiency by compressing the discrete tokens in explicit CoT into a few continuous steps.
For example, iCoT-KD~\citep{deng2023implicit, deng2024explicit} distills entire reasoning path into a single inference step,
%~(a single time step in LLM inference) before generating the final answer, 
while curriculum learning is adopted in iCoT-SI~\citep{deng2024explicit} to gradually encourage the model to shift from explicit rationales toward fully latent reasoning.
However, compressing an entire CoT reasoning path, which may consist of thousands of tokens, into a single inference step, is too aggressive and degrades model performance. 
Consequently, more recent works adopt intermediate continuous vectors, each representing a block of consecutive discrete tokens in CoT reasoning.
For example, Coconut~\citep{hao2024training} progressively replaces blocks of discrete word tokens with continuous vectors.
%and concatenates them together in the reasoning chain. 
CoLaR~\citep{tan2025think}, CODI~\citep{shen2025codi}, SIM-CoT~\citep{wei2025sim}, and CoT2~\citep{zhang2025cot2} further add supervision to align the latent states with token blocks they replace in explicit CoT, with CoT2 forming continuous tokens as convex combinations of vocabulary embeddings.
While improving inference efficiency, these methods essentially remain a form of token compression and could still be restricted by token-level reasoning.
Moreover, their continuous vectors are updated either in a fixed manner or via simple Gaussian transitions, which limits their ability to model the diverse and exploratory thinking.
\begin{figure*}[t]
\centering
\includegraphics[width=0.76\linewidth]{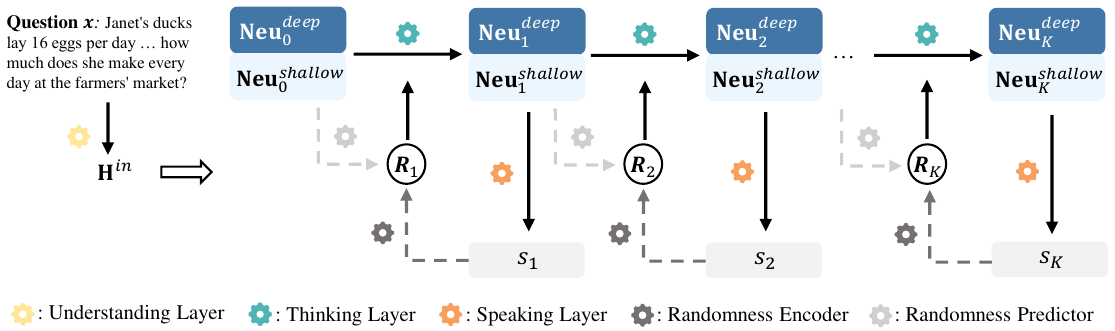}
\caption{Illustration of our {\ModelName} model. In the thinking stage, {\ModelName} reasons in the continuous space for $K$ steps by updating neurons $\mathbf{Neu}_k^{deep}$ and 
$\mathbf{Neu}_k^{shallow}$.
%, which can be decoded in parallel to natural language in the speaking stage. 
In the speaking stage, $\{\mathbf{Neu}_k^{shallow},k=1,...,K\}$ are translated to natural language in parallel. If intermediate steps are not required, we only need to translate $\mathbf{Neu}^{shallow}_K$ to a final answer. 
During training, the random variable $\bm{R}_k$ is derived from the ground-truth sentence 
$\bm{s}_{k}$ using the randomness encoder $f$, while in inference, it is sampled from a distribution predicted by the randomness predictor $g$ based on $\mathbf{Neu}^{deep}_k$, $\mathbf{Neu}^{shallow}_k$, and $\text{H}^{in}$.
% \textcolor{blue}{Probably shouldn't hide the blue dashed line behind Neu.? Probably we should replace Neu with N, which is more concise}
}
\label{figure_framework}
\vspace{-9pt}
\end{figure*}
\textbf{Depth-axis latent reasoning.}
These approaches modify the transformer architecture to support latent reasoning. One representative is the Looped Transformer~\citep{giannou2023looped,yanglooped}, which 
%introduces recurrence within transformer layers, 
reuses the same layer parameters across iterations to deepen internal deliberation. For example, ~\citet{geiping2025scaling} apply transformer blocks recurrently for each token,
%. It allows the model to unroll to arbitrary depth, 
enabling test-time scaling up without generating additional tokens.~\citet{zhu2025scaling} further incorporate adaptive early-exit mechanisms, allowing the model to dynamically determine the number of recurrent steps.
Compared with \ModelName, these methods lack supervision for intermediate reasoning states.
Consequently, they can be seen as parameter-efficient ways to increase the depths of transformers, which are orthogonal to our focus.
%gives limited performance gain and significantly slows down reasoning. Therefore, they are not the focus of our study.
%One early effort in this direction is Implicit CoT (iCoT)~\citep{deng2023implicit}, which transfers the reasoning representation of a CoT-enabled teacher to a student model through knowledge distillation. Building on this,~\cite{deng2024explicit} further introduced a training strategy that gradually reduces the supervision of intermediate steps, encouraging the model to progressively shift from relying on external rationales to performing reasoning as an internal latent process. Recently, Coconut~\citep{hao2024training} advanced this direction by concatenating hidden states with word tokens in the reasoning chain, effectively allowing the model to switch between ``language mode'' and ``latent mode''. Another representative work is CoLaR~\citep{tan2025think}, which compresses reasoning chains into a dense latent space via a two-stage training approach, allowing the model to reason silently and dynamically adjust reasoning speed at inference. 
\section{Our {\ModelName} Model}

In this section, we introduce {\ModelName}, a new architecture for fully continuous, stepwise reasoning at the thought level. 
Specifically, we model the reasoning process as a conditional hidden Markov model~(cHMM), where the input questions are the conditions, the internal thoughts are hidden variables, and spoken sentences are observable variables. 
This formulation enforces a clear disentanglement between thinking and speaking: thinking corresponds to transitions of continuous thoughts, and speaking is the emission process from these thoughts to natural language. Crucially, continuous thought at each time step represents a \emph{reasoning step}, which enables the model
to plan thoroughly before producing a sentence. Compared with traditional transformers, our paradigm offers two key
advantages: (1) by disentangling thinking from speaking,
it enables more structured and deliberate reasoning beyond
token-by-token generation; and (2) by carrying out the entire thinking process in continuous space, we significantly increase the information bandwidth between reasoning steps, which enables parallel thinking within a single forward pass of {\ModelName}.
%the entire thinking process is carried out in continuous space, thereby supporting
%higher information bandwidth and parallel thinking
% This enables the model to plan thoroughly before saying a single word. 
% Besides, our paradigm offers another advantage compared with traditional transformers: the entire thinking process is now carried out in a continuous space.
% By removing the discretization operation in information flow, we significantly increase the information bandwidth between steps, which even enables parallel thinking in a single forward pass of {\ModelName}.

%Compared with traditional transformers, our paradigm offers two key advantages: (1) by disentangling thinking from speaking, it enables more structured and deliberate reasoning beyond token-by-token generation; and (2) the entire thinking process is carried out in continuous space, thereby supporting higher information bandwidth and parallel thinking.

% In {\ModelName}, the continuous thoughts are represented by the states of a group of neurons, and thinking is essentially mapping the current neuron states to the next reasoning steps.
For the transition between thoughts, the simplest idea is to have a deterministic mapping from the current thought to the next one.
%between continuous tokens are typically static mappings in most previous work~\cite{hao2024training,shen2025codi}.
However, this could be suboptimal, since there are usually multiple plausible reasoning paths for a math problem, which makes deterministic mappings insufficient to model the diverse distribution of possible thoughts, 
limiting model exploration during inference. 

To tackle this problem, {\ModelName} maps each thought to a multi-modal distribution of possible thoughts at the next step.
By doing so, we can explicitly parameterize the randomness in the thinking process without relying on token-level sampling. 

We will first introduce the structure of {\ModelName} in Section~\ref{section:model_structure}. 
Then, we will show the training recipe of {\ModelName} in Section~\ref{section:training}.
Specifically, we propose a two-phase training scheme, inspired by VAE~\citep{kingma2013auto}, to address the intractability of sample probabilities caused by the stochastic mapping between latent thoughts. %Due to page limit, we leave realization details in Section~\ref{append:eva}.

%the stochastic mapping poses a challenge in training: under the complicated transition, the probability of a ground-truth step becomes intractable to maximize directly. 
%To address this, inspired by VAE~\citep{kingma2013auto}, we propose a two-phase training scheme that infers the posterior distribution of thoughts given the ground-truth reasoning step and optimizes an ELBO-like objective. 
%To simplify training and injectinductive biases for more interpretable control over randomness, we further incorporate ideas from sparse autoencoders~\citep{cunningham2023sparse}, which we will elaborate on in Section~\ref{section:training}.

\subsection{Model Architecture}\label{section:model_structure}
This section explains how {\ModelName} ``thinks'' and generates the solution to a question. As shown in Figure~\ref{figure_framework}, it consists of three separate layers: understanding, thinking, and speaking, which provide conditions, transition probabilities, and emission probabilities for cHMM.

\textbf{Understanding.}
Given the input question $x$ of length $n$, we use an understander layer to convert it into a sequence of feature vectors $\text{H}^{in} = \{\bm{h}_1,...,\bm{h}_n\}$:
\begin{equation}
    \text{H}^{in} = \text{Understanding}(x) \in \mathbb{R}^{n\times d},
\end{equation}
where $d$ denotes the latent dimension.

\textbf{Thinking.} As the hidden variable of the cHMM, our continuous thought is realized as the states of a group of neurons. Specifically, inspired by neuroscience evidence of functional specialization in the human brain~\citep{kanwisher2010functional}, we have two types of neurons: $\mathbf{Neu}^{deep}\in \mathbb{R}^{T\times d}$ and $\mathbf{Neu}^{shallow}\in \mathbb{R}^{S\times d}$, in charge of slow and fast thinking, respectively.
The intuition is that $\mathbf{Neu}^{deep}$ facilitates deeper reasoning, while $\mathbf{Neu}^{shallow}$ represents shallower and more readily verbalizable reasoning, as it interacts more closely with the speaking layer (see Eq.~\ref{eq:speaker}). 
Here, $T$ and $S$ denote the numbers of neurons in each group. 

Then, each thinking step corresponds to an update of $\mathbf{Neu}^{deep}$ and $\mathbf{Neu}^{shallow}$ within a thinking layer, conditioned on the input feature $\text{H}^{in}$ and their own previous states. 

As described earlier, a key difference between {\ModelName} and existing continuous reasoning approaches is that it explicitly parameterizes the randomness of the thinking process at the \emph{step level}. 
Concretely, we introduce an auxiliary random variable $\bm{R}_k\in \mathbb{R}^{\tau \times d}$ to control the reasoning behavior at step $k$. $\tau$ here is the length scaling factor of $\bm{R}_k$.
%, which is usually smaller than the total numbers of neurons.
Instead of directly mapping the current thought at time step $k-1$ to the multi-modal and high-dimensional distribution of thoughts at the next step $k$, we first map it to a distribution of $\bm{R}_k$, which lies in a better-behaved, lower-dimensional space.
Then, we can map the current thought and the controlling variable $\bm{R}_k$ deterministically to the next thought.
To encourage each dimension of $\bm{R}_k$ to represent a single random factor and stabilize the training process, we further add a sparsity constraint to $\bm{R}_k$, which will be elaborated in Section~\ref{section:training}. As observed in Sections~\ref{sec:syn_results} and~\ref{section:analysis}, the introduction of $\bm{R}_k$ facilitates more controllable randomness in the reasoning process. 
For example, we find that different dimensions of $\bm{R}_k$ effectively controls different aspects of thinking, including high-level factors (e.g., search directions, reasoning depth) and low-level choices (e.g., expression format, sentence length). 

Specifically, we employ a learnable randomness predictor $g$ to predict the distribution of $\bm{R}_k$ based on input information $\text{H}^{in}$ and the current thought $\mathbf{Neu}_{k-1}^{deep}$ and  $\mathbf{Neu}_{k-1}^{shallow}$:
\begin{align}
    \bm{\mu}_{k}, \bm{\sigma}_{k} &= g(\mathbf{Neu}_{k-1}^{deep},  \mathbf{Neu}_{k-1}^{shallow}, \text{H}^{in})\label{eq_ranpre}.
\end{align}
Then, a sample of $\bm{R}_k\sim\mathcal{N}(\bm{\mu}_{k}, \bm{\sigma}_{k})$ is drawn to guide the transition from the current thought to the next one\footnote{Eq.~\eqref{equ:thinker} can be extended to include all preceding thinking neurons as input. In Appendix~\ref{append:longrange}, we show that this first-order setting does not create an information bottleneck compared to CoT.}:
\begin{equation}
\begin{aligned}\label{equ:thinker}
    \mathbf{Neu}_{k}^{deep},  \mathbf{Neu}_{k}^{shallow}= &\text{Thinking}(\mathbf{Neu}_{k-1}^{deep},  \mathbf{Neu}_{k-1}^{shallow}, \\ & \text{H}^{in}, \bm{R}_k).
\end{aligned}
\end{equation}
The thinking layer uses bi-directional attention to facilitate this information flow and iteratively updates the neurons to stimulate step-by-step reasoning. 
The initial values of neurons, $\mathbf{Neu}_{0}^{deep}$ and  $\mathbf{Neu}_{0}^{shallow}$, are learnable parameters.

\textbf{Speaking.} After each thinking step, some thoughts in $\mathbf{Neu}_{k}^{shallow}$, such as plans and intermediate conclusions, may become ready for verbalization. We use a speaking layer to translate them into the discrete space of sentences:
\begin{equation}\label{eq:speaker}
    \bm{s}_k = \text{Speaking}(\mathbf{Neu}^{shallow}_k).
\end{equation}
Unlike traditional transformers, thinking in {\ModelName} does not depend on speaking, so our speaking stage is optional for all intermediate steps and can be skipped except for the final step when only an answer is required. 
Moreover, speaking at each step is independent to each other, making them highly parallelizable compared with traditional models.

In addition, since randomness in {\ModelName} is resolved at the step level in the thinking stage \emph{before} the speaking layer generates specific sentences, complex randomness can be simulated ``in one pass'' rather than through multiple ``iterative'' sampling. %there is little randomness left in the actual generation of tokens, which is different from traditional transformers. 
This provides a new perspective for the development of non-autoregressive (NAR) language models with one-pass generation, and suggests a promising avenue for accelerating diffusion LLMs~\citep{gulrajani2023likelihood,sahoo2024simple}. 
To verify this idea, 
we performed a preliminary experiment by replacing the speaking layer of {\ModelName} with a fully NAR decoder. 
As shown in Section~\ref{section:analysis}, even the simplest NAR decoding strategy still obtains competitive performance, which provides further acceleration to \ModelName. 
This also highlights that our model achieves a clear separation between thinking and speaking, ensuring that the reasoning quality remains stable while allowing flexibility in the choice of speaking strategy.

\subsection{Training}\label{section:training}

In this part, we discuss how to effectively train {\ModelName}.
Assume we have a training sample that consists of a question $x$ and its step-by-step solution $y
=\{y_1, ..., y_N\}$.
The training objective is to find the parameters that maximize the probability of generating all $y_k$.
However, because sampling occurs in thinking rather than in speaking, we no longer have a closed-form conditional probability for $y_k$.

To solve this, we adapt the evidence lower bound~(ELBO) used in VAE training to maximize the marginal probability of $y_k$ at each step. 
Specifically, the general ELBO is:
\begin{equation}
\small
\mathcal{L}^{\text{ELBO}}_k = 
\mathbb{E}_{q(\bm{R}_k|y_k)} \big[ \log p(y_k|\bm{R}_k) \big] 
- \mathrm{KL} \big( q(\bm{R}_k|y_k) \,\|\, p(\bm{R}_k) \big),
\end{equation}
which is a combination of a reconstruction term and a KL term.
In our case, the VAE encoder $q$ is realized by a learnable function $f$ (referred to as randomness encoder in Figure~\ref{figure_framework}) to map $y_k$ to the posterior mean and variance of $\bm{R}_k$:
\begin{equation}
\bm{\hat{\mu}}_{k}, \bm{\hat{\sigma}}_{k} =f(y_k).
\label{eq:encoder}
\end{equation}
The VAE decoding is simply a step of thinking and speaking in Eqs.~\ref{equ:thinker} and~\ref{eq:speaker}, which maps $\bm{R}_k$ back to the sentence space.
Then the reconstruction loss can be expanded as:
\begin{equation}
\begin{aligned}
    \mathcal{L}^{re}_k =&\mathbb{E}_{\bm{R}_k\sim \mathcal{N}(f(y_k))} \big[ \log p(\bm{s}_k=y_k|\mathbf{Neu}_{k-1}^{deep}, \\ & \mathbf{Neu}_{k-1}^{shallow}, 
    \text{H}^{in}, \bm{R}_k) \big].
\end{aligned}
\end{equation}

Unlike vanilla VAEs with fixed isotropic Gaussian priors, our randomness predictor $g$ in Eq.~\ref{eq_ranpre} provides a different prior for different questions and thoughts as described in Section~\ref{section:model_structure}. 
Consequently, our KL divergence between the prior (Eq.~\ref{eq_ranpre}) and the posterior (Eq.~\ref{eq:encoder}) is:
\begin{equation}
\small
    \mathcal{L}^{KL}_k=\mathrm{KL} \big(\mathcal{N}(f(y_k))  \,\|\, \mathcal{N}(g(\mathbf{Neu}_{k-1}^{deep},  \mathbf{Neu}_{k-1}^{shallow}, \text{H}^{in})) \big).
\end{equation}

In addition, without a proper control, the model may encode excessive semantic information of $y_k$ into $\bm{R}_k$ via the randomness encoder $f$. This information can be easily merged into $\mathbf{Neu}_{k}^{shallow}$ in Eq.\ref{equ:thinker} for the reconstruction of $y_k$, creating a training ``shortcut'' that prevents the model from learning genuine reasoning. To prevent this, we introduce an extra sparsity loss $\mathcal{L}^{sparse}_k = \|\bm{R}_k\|_1$. This forces $\bm{R}_k$ to store only the minimal amount of information that cannot be inferred from context, while providing a stronger inductive bias for structured and interpretable reasoning. For example, at each thinking step, only a small set of dimensions need to be active. Dimensions useful for breadth-first reasoning may not be activated during depth-first reasoning. 
To reflect this inductive bias and encourage each dimension of $\bm{R}_k$ to capture a ``mono-semantic'' factor in thinking, we draw inspiration from sparse autoencoders and implement the function $f$ as a $d$-dimensional transformer followed by an MLP that expands the dimension from $d$ to $\tau\times d$. The overall training loss of {\ModelName} is:
\begin{equation}\label{eq:loss}
    \mathcal{L}= \sum_{k\in \{1,2,...,N\}} \big[- \mathcal{L}_k^{re} + \mathcal{L}_k^{KL} + \lambda \mathcal{L}^{sparse}_k\big],
\end{equation}
where $\lambda$ is a hyperparameter that controls the strength of the sparsity loss. To eliminate the need to pre-select $\lambda$, we adopt a dynamic weighting strategy that adaptively controls $\lambda$ to balance the reconstruction and sparsity regularization during training (see Appendix~\ref{append:dynamic}). 
Moreover, to avoid model collapse caused by simultaneously tuning the posterior and prior, we adopt a two-phase training scheme, where we first optimize $\mathcal{L}_k^{re}$ and $\mathcal{L}^{sparse}_k$ with fixed $g$ and then train $g$ to minimize $\mathcal{L}_k^{KL}$ with all other parameters fixed.

\section{Experiments}
\label{experiments}
\subsection{Evaluation on Synthetic Benchmarks}
\begin{figure}[t]
\centering
\includegraphics[width=0.9\linewidth]{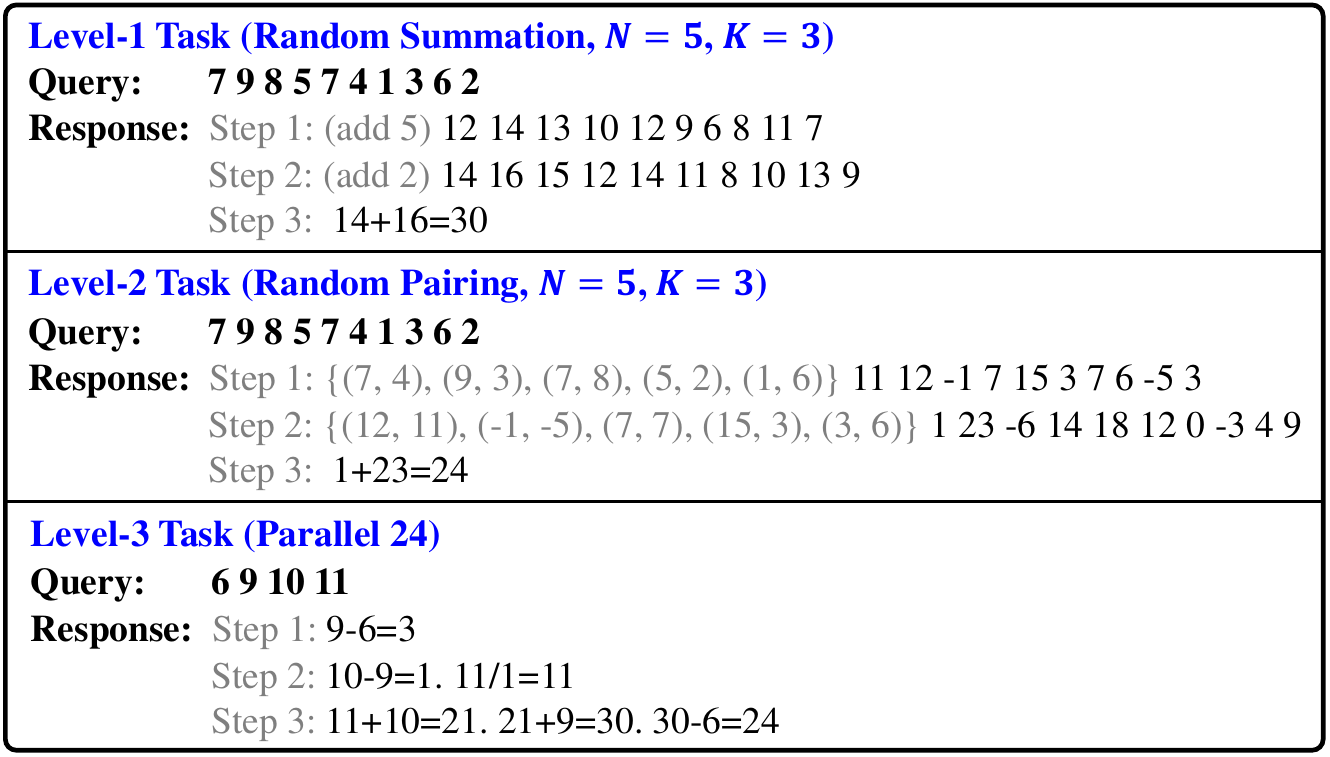}
\caption{Overview of synthetic tasks. The output sentence is shown in black, while the gray segments are added only to facilitate step delineation and to understand the reasoning process.
}
\label{task_all}
\vspace{-5pt}
\end{figure}

To systematically compare the fundamental characteristics of {\ModelName} and transformers in performing reasoning and randomness modeling, we first design three synthetic reasoning tasks, with examples provided in Figure~\ref{task_all}.
%before moving to large-scale evaluations on benchmarks.
%: \textbf{Pairwise Calculation} and \textbf{Parallel 24}. These tasks are intended to isolate key challenges in stochastic reasoning, including long-range thought maintenance, step-level randomness modeling, and the decoupling between contextual memory and stochastic transitions.
\subsubsection{Experimental Setup}
\textbf{Level-1 Task: Random Summation (RS).} Starting from a random sequence of $2N$ 1-digit non-negative integers, at each step, another 1-digit non-negative integer is uniformly sampled and added to each number in the sequence. 
This process is repeated for $K$ steps. In the last step, the sum of the first two numbers is computed as the final answer. 
This task requires the most basic arithmetic skills and modeling of very simple randomness in the addends.
%the least token-level randomness, because randomness can be resolved at the first generated token.
%and the entire reasoning trajectory becomes conditionally deterministic once the initial random integer is sampled.
% \textbf{Motivation.} This task is designed for comparison with RP task. In this task, the randomness can be resolved at the first generated token. As a result, the entire reasoning becomes conditionally deterministic given this initial random choice, which is suitable for token-based autoregressive modeling. 

\textbf{Level-2 Task: Random Pairing (RP).} Given a random sequence of $2N$ 1-digit non-negative integers, at each step, the model needs to randomly partition it into $N$ pairs. 
For each pair $(a,b)$, the first position is updated to $(a+b)$ and the second position to $(a-b)$. The other settings are the same as in the level-1 RS task. Compared with RS, this task not only involves moderately more complex calculations, but also induces more complicated randomness of partitioning.
Because a global partitioning configuration must be determined before token generation, the randomness is more difficult to be modeled via token-level sampling.
%because, before token generation, the pairing configuration must be randomly selected from all possible pairings. Such randomness is difficult to model via token-level sampling, as it requires committing to a global discrete pairing structure rather than making local token-wise decisions. 
% \begin{figure}[h]
% \centering
% %\setlength{\abovecaptionskip}{2pt}
% \includegraphics[width=\linewidth]{figures/syn_figure.pdf}
% \label{task_1}
% \vspace{-9pt}
% \end{figure}
% \begin{figure}[h]
% \centering
% %\setlength{\abovecaptionskip}{2pt}
% \includegraphics[width=\linewidth]{figures/syn_figure2.pdf}
% \label{task_2}
% \vspace{-9pt}
% \end{figure}

\textbf{Level-3 Task: Parallel 24 (P24).} The classic Game of 24 requires a model to perform trial and error until reaching 24. In P24, we ask models to think about different paths in parallel and find a correct one in $K$ steps. 
%At each intermediate step, models are required to output the most promising incomplete path, but the next 
%Unlike standard formulations, given four numbers (e.g., 6, 9, 10, 11), at each step the model may reconsider all numbers from all prior reasoning paths. For example, it may start with $9-6=3$, then reselect and compute $10-9=1$, $11/1=11$, and finally replan to form $11+10=21$, $21+9=30$, $30-6=24$. Each step allows a fresh choice over all operands and operations. 
This task requires parallel thinking and a larger information bandwidth, as the model needs to consider as many reasoning paths as possible at each step. 
It simulates how humans dynamically adjust their reasoning strategy during problem-solving.
%Therefore, it provides a direct testbed for contrasting the advantages of continuous reasoning over token-based reasoning.
% \begin{figure}[h]
% \centering
% %\setlength{\abovecaptionskip}{2pt}
% \includegraphics[width=\linewidth]{figures/syn_figure3.pdf}
% \label{task_3}
% \vspace{-9pt}
% \end{figure}
\subsubsection{Results}\label{sec:syn_results}
\begin{table}[!t]
\setlength{\abovecaptionskip}{5pt}
\centering
\small
\caption{Accuracy on synthetic tasks ($N=5,K=3$).}
\begin{tabular}{l|p{1.2cm} p{1.2cm} p{1.2cm}}
\toprule[1.5pt]
Model    &   RS & RP &  P24 \\
\midrule[1.2pt]
Transformer & 0.999  & 0.589   & 0.170 \\
%CoT (3B)    & 0.999 & 0.764    & 0.183 \\
{\ModelName} &  1.000 & 0.712   & 0.668 \\
\bottomrule[1.5pt]
\end{tabular}
\vspace{-8pt}
\label{tab:syn}
\end{table}
\begin{figure}[t]
\centering
\setlength{\abovecaptionskip}{2pt}
\includegraphics[width=1\linewidth]{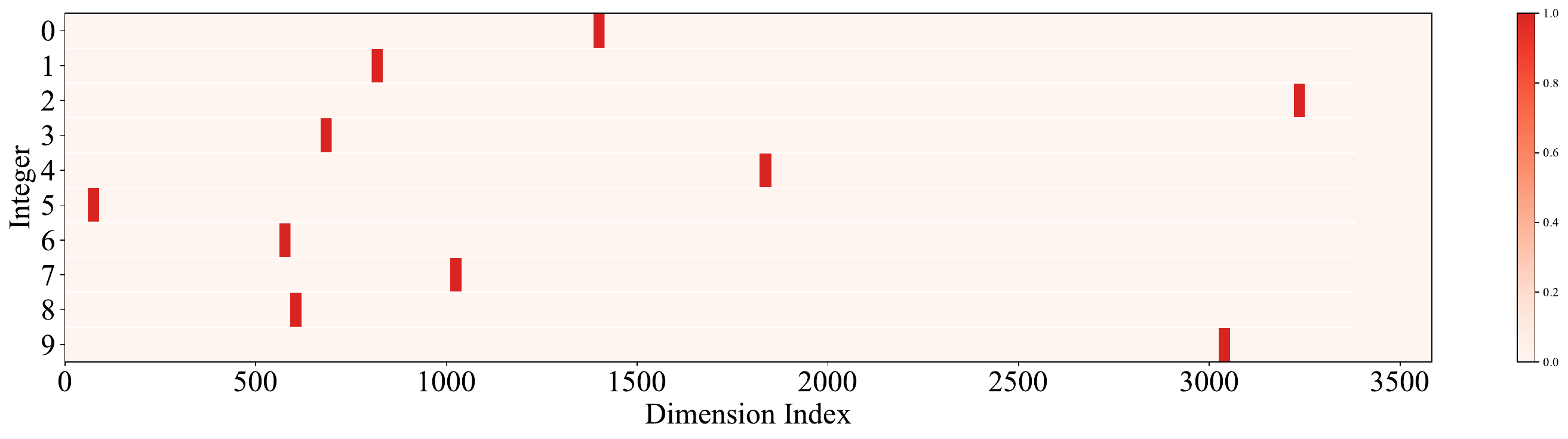}
\caption{Emergent one-to-one encoding of randomness in $\bm{R}_k$. Each randomly chosen
integer consistently activates a unique dimension in $\bm{R}_k$, regardless of the input sequence.
}
\label{syn_tr}
\vspace{-15pt}
\end{figure}
\begin{table*}[t]
\caption{
Results of models \textbf{trained from scratch}. All continuous reasoning baselines (0.5B) are trained on equation data, following their original papers. We report the best continuous models in bold and the runner-ups with underline.
}
\centering
\small
\setlength{\tabcolsep}{7pt}
\renewcommand{\arraystretch}{1.}
\begin{tabular}{l|l|ll|ll|ll}
\toprule[1.5pt]
 & & \multicolumn{2}{c|}{GSM8K} & \multicolumn{2}{c|}{SVAMP} & \multicolumn{2}{c}{MultiArith} \\
\cmidrule(lr){3-4} \cmidrule(lr){5-6} \cmidrule(lr){7-8}
Model & Train (h) & Acc (\%) & Test (s)  & Acc (\%) & Test (s)  & Acc (\%) & Test (s) \\
\midrule[1.2pt]
\multicolumn{8}{c}{\textit{\textbf{Training Data: Text}}} \\
\midrule
CoT-SFT & ${0.31 \scriptstyle \pm 0.01}$ & ${13.27 \scriptstyle \pm 0.59}$  & ${80.95 \scriptstyle \pm 5.78}$ & ${23.25 \scriptstyle \pm 0.39}$ & ${216.56 \scriptstyle \pm 11.88}$ & ${25.20 \scriptstyle \pm 0.47}$ & ${18.32 \scriptstyle \pm 1.81}$ \\
\midrule
\cellcolor{gray!15}{\ModelName} & \cellcolor{gray!15}${1.13 \scriptstyle \pm 0.01}$ & \cellcolor{gray!15}${\textbf{17.97} \scriptstyle \pm 0.96}$ & \cellcolor{gray!15}${17.53 \scriptstyle \pm 0.29}$ & \cellcolor{gray!15}${\textbf{24.39} \scriptstyle \pm 1.10}$ & \cellcolor{gray!15}${34.14 \scriptstyle \pm 0.49}$ & \cellcolor{gray!15}${\textbf{23.67} \scriptstyle \pm 0.53}$ & \cellcolor{gray!15}${5.90 \scriptstyle \pm 0.17}$ \\
$-\ w/o\ \mathbf{Neu}^{deep}$ &${1.12 \scriptstyle \pm 0.01}$ & ${16.91 \scriptstyle \pm 1.60}$ & ${17.28 \scriptstyle \pm 0.13}$ & ${22.04 \scriptstyle \pm 0.29}$ & ${34.83 \scriptstyle \pm 0.26}$ & ${20.28 \scriptstyle \pm 0.25}$ & ${5.68 \scriptstyle \pm 0.16}$ \\
$-\ w/o\ \bm{R}_k$ & ${0.77 \scriptstyle \pm 0.01}$ & ${16.07 \scriptstyle \pm 0.00}$ & ${17.20 \scriptstyle \pm 0.10}$ & ${20.27 \scriptstyle \pm 0.00}$ & ${32.43 \scriptstyle \pm 0.11}$ & ${16.66 \scriptstyle \pm 0.00}$ & ${5.65 \scriptstyle \pm 0.12}$ \\
%$-\ w/o\ sparsity$ & ${1.14 \scriptstyle \pm 0.01}$ & ${1.16 \scriptstyle \pm 0.09}$ & ${16.99 \scriptstyle \pm 0.10}$ & ${2.16 \scriptstyle \pm 0.09}$ & ${32.35 \scriptstyle \pm 2.01}$ & ${0.78 \scriptstyle \pm 0.09}$ & ${6.15 \scriptstyle \pm 0.21}$ \\
\midrule[1.2pt]
\multicolumn{8}{c}{\textit{\textbf{Training Data: Equation}}} \\
\midrule
CoT-SFT & ${0.38 \scriptstyle \pm 0.01}$ & ${20.31 \scriptstyle \pm 0.85}$  & ${32.97 \scriptstyle \pm 1.03}$ & ${27.92 \scriptstyle \pm 0.36}$ & ${162.12 \scriptstyle \pm 7.43}$ & ${41.48 \scriptstyle \pm 0.50}$ & ${9.61 \scriptstyle \pm 0.91}$\\
iCoT-SI & ${0.43 \scriptstyle \pm 0.03}$ & ${14.81 \scriptstyle \pm 0.55}$  & ${13.11 \scriptstyle \pm 1.62}$ & ${18.46 \scriptstyle \pm 0.54}$ & ${17.30 \scriptstyle \pm 0.31}$ & ${21.33 \scriptstyle \pm 0.43}$ & ${2.32 \scriptstyle \pm 0.05}$ \\
Coconut& ${1.50 \scriptstyle \pm 0.25}$ & ${15.27 \scriptstyle \pm 0.05}$  & ${59.53 \scriptstyle \pm 0.53}$ & ${17.76 \scriptstyle \pm 0.01}$ & ${106.74 \scriptstyle \pm 1.46}$ & ${15.50 \scriptstyle \pm 0.01}$ & ${15.98 \scriptstyle \pm 0.28}$  \\
CoLaR & ${1.17 \scriptstyle \pm 0.00}$ & ${15.45 \scriptstyle \pm 0.42}$  & ${17.20 \scriptstyle \pm 0.45}$ & ${23.09 \scriptstyle \pm 0.09}$ & ${35.80 \scriptstyle \pm 0.45}$ & ${38.60 \scriptstyle \pm 0.55}$ & ${3.80 \scriptstyle \pm 0.45}$ \\
CODI & ${0.68 \scriptstyle \pm 0.09}$ & ${\underline{16.07} \scriptstyle \pm 0.17}$ & ${8.58 \scriptstyle \pm 0.25}$ & ${\underline{26.24} \scriptstyle \pm 0.22}$ & ${24.88 \scriptstyle \pm 0.26}$ & ${\underline{40.03} \scriptstyle \pm 0.25}$ & ${3.48 \scriptstyle \pm 0.24}$ \\
\midrule
\cellcolor{gray!15}{\ModelName} & \cellcolor{gray!15}${1.03 \scriptstyle \pm 0.01}$ & \cellcolor{gray!15}${\textbf{24.11} \scriptstyle \pm 0.97}$ &\cellcolor{gray!15} ${17.76 \scriptstyle \pm 0.04}$ & \cellcolor{gray!15}${\textbf{27.77} \scriptstyle \pm 0.32}$ & \cellcolor{gray!15}${34.02 \scriptstyle \pm 0.11}$ & \cellcolor{gray!15}${\textbf{42.33} \scriptstyle \pm 0.56}$ & \cellcolor{gray!15}${6.02 \scriptstyle \pm 0.28}$ \\
$-\ w/o\ \mathbf{Neu}^{deep}$ &${0.99 \scriptstyle \pm 0.01}$ & ${21.07 \scriptstyle \pm 0.76}$ & ${17.09 \scriptstyle \pm 0.30}$ & ${25.83 \scriptstyle \pm 0.19}$ & ${35.24 \scriptstyle \pm 0.11}$ & ${38.38 \scriptstyle \pm 1.04}$ & ${6.00 \scriptstyle \pm 0.29}$ \\
$-\ w/o\ \bm{R}_k$ & ${0.74 \scriptstyle \pm 0.01}$ &${23.99 \scriptstyle \pm 0.00}$ & ${17.67 \scriptstyle \pm 0.09}$ & ${26.90 \scriptstyle \pm 0.00}$ & ${33.78 \scriptstyle \pm 0.54}$ & ${42.17 \scriptstyle \pm 0.00}$ & ${5.99 \scriptstyle \pm 0.08}$ \\
%$-\ w/o\ sparsity$ & ${1.01 \scriptstyle \pm 0.02}$ & ${1.16 \scriptstyle \pm 0.10}$ & ${17.00 \scriptstyle \pm 0.21}$ & ${1.33 \scriptstyle \pm 0.05}$ & ${27.63 \scriptstyle \pm 2.08}$ & ${0.56 \scriptstyle \pm 0.10}$ & ${4.96 \scriptstyle \pm 2.57}$ \\
\bottomrule[1.5pt]
\end{tabular}
\label{tabel:main}
\vspace{-5pt}
\end{table*}
Table~\ref{tab:syn} shows the performance of {\ModelName} and traditional transformer on all three tasks. 
On the RS task, where the randomness can be effectively absorbed into a sequential token-generation process, both {\ModelName} and the transformer achieve perfect performance. 
However, the performance of transformer degrades sharply on the RP task, which requires the modeling of the more structured stochasticity of random partitioning. 
This result suggests that token-based sampling is better suited to randomness that admits a sequential factorization, but becomes severely limited when dealing with structured randomness that involves global thinking.
%\textcolor{red}{Ning: we should talk about how Marcos perform on RP, and explain a little bit.}

On the more challenging P24 task, {\ModelName} achieves a substantial improvement compared with transformer. This large gap highlights the advantage of high-bandwidth continuous thinking in our thinking layer. 
By implicitly maintaining all parallel reasoning paths, {\ModelName} explores the solution space far more effectively than transformer, which is important in settings with heavy branching and frequent replanning.

An interesting phenomenon is that our model learns to store perfectly clean, context-independent randomness in the random variable $\bm{R}_k$. 
For example, on the RS task, each randomly chosen integer consistently activates a unique dimension of $\bm{R}_k$, while leaving all other dimensions inactive (Figure~\ref{syn_tr}). 
This pattern holds across ALL inputs. More importantly, during inference, if we fix $\mathbf{Neu}_{k-1}^{deep},  \mathbf{Neu}_{k-1}^{shallow}$ and manually modify $\bm{R}_k$ at a given step $k$, the decoded
$\bm{s}_k$ and all subsequent reasoning steps change consistently according to this
modification, without any error (accuracy = 100\%). 
These findings demonstrate that {\ModelName} explicitly separates randomness from contextual processing in a structured and interpretable manner. The thinking neurons are responsible for processing contextual information and performing the core reasoning computations, while $\bm{R}_k$ controls the direction of reasoning (i.e., the randomness). 

\subsection{Evaluation on Real-World Benchmarks}
Having validated the efficacy of our {\ModelName} on synthetic tasks, we now turn to larger-scale real-world tasks to further evaluate its potential on more challenging problems.
\subsubsection{Experimental Setup}\label{section:exp_setup}
% \textcolor{red}{Ning: One problem is that our parameter size is 3x more compared with other models? How should to justify the fairness of the comparison. Maybe the same or lower tflops?}
% \textcolor{red}{Ning: We may also want to show the training wall-clock time or flops if our results looks better.}
\textbf{Datasets and Tasks.} Following~\cite{tan2025think}, we train our model on \textbf{GSM8K-Aug} dataset~\citep{deng2023implicit},
%, an augmented version of GSM8K~\citep{cobbe2021training}, 
which contains 385K training samples. Then, we evaluate on the original GSM8K test set (in-domain task), as well as two out-of-domain benchmarks: (1) \textbf{SVAMP}~\citep{patel2021nlp}, consisting of 4,138 arithmetic problems constructed by subtle variations in wording and semantic perturbations, which aims to evaluate the robustness of reasoning, (2) \textbf{MultiArith}~\citep{roy2015solving}, a collection of 600 problems from MAWPS~\citep{koncel2016mawps} that require multi-step reasoning. Our evaluation metrics include (1) Accuracy (Acc), which reflects the reasoning ability; (2) Training Time (Train), which measures the time cost for model training; and (3) Test Time (Test), which measures inference efficiency by counting the total time required to produce solutions for all test samples. More detailed descriptions of the evaluation setup are provided in Appendix~\ref{append:eva}.

\textbf{Baselines.} We compare with (1) \textbf{CoT}~\citep{wei2022chain}, which trains transformer-based models on problem-solution pairs with explicit reasoning paths, 
(2) \textbf{iCoT-SI}~\citep{deng2024explicit}, which gradually removes explicit reasoning steps during training to encourage implicit reasoning, 
(3) \textbf{Coconut}~\citep{hao2024training}, which treats latent representations as a special token and inputs them alongside word tokens into an autoregressive model, 
(4) \textbf{CoLaR}~\citep{tan2025think}, which compresses reasoning tokens and learns to predict the distribution of these compressed embeddings, and 
(5) \textbf{CODI}~\citep{shen2025codi}, which self-distills hidden activations across all layers at the selected distillation token.
\textbf{Implementation Details.} 
We implement the understanding, thinking, speaking layers and randomness encoder as 0.5B models (structure detailed in Appendix~\ref{append:eva}). The randomness predictor is a two-layer MLP. 
%Therefore, the total parameters of {\ModelName} are approximately $0.5\text{B}\times4 = 2\text{B}$. 
For fair comparison, we implement baselines with both 0.5B and 3B parameters. For our model and CoT-SFT, we include both an equation-based version for direct comparison and a text-based version for real-world applications.
For other baselines, we follow their original recipes to first train an explicit CoT for initialization and then finetune with only structured equations. 

\begin{table}[t]
\centering
% \small
\caption{Results of models initialized from pretrained checkpoints. For {\ModelName}, we remove $\bm{R}_k$ since pretrained weights are not naturally suited for its stochastic role.}
\label{tab:combined_results}
\begin{tabular}{lccc}
\toprule[1.5pt]
Model & GSM8K & SVAMP & MultiArith \\
\midrule[1.2pt]
\multicolumn{4}{c}{\textit{Backbone: Qwen2.5-0.5B}} \\
\midrule
CoT & \underline{41.6} & \underline{71.9} & \underline{88.3} \\
Coconut & 28.6 & 60.7 & 88.0 \\
CoLaR & 36.0 & 61.7 & 82.2 \\
CODI & 40.8 & 71.2 & 87.6 \\
\rowcolor{gray!15}{\ModelName ($w/o\ \bm{R}_k$)} & \textbf{43.8} & \textbf{75.1} & \textbf{98.3} \\
\midrule[1.2pt]
\multicolumn{4}{c}{\textit{Backbone: LLaMA-3.2-1B-Instruct}} \\
\midrule
CoT & \textbf{61.6} & \underline{66.7} & \textbf{99.3} \\
Coconut & 45.3 & 48.8 & 90.1 \\
CoLaR & 40.1 & 54.9 & 96.1 \\
CODI & 55.6 & 61.1 & 91.3 \\
\rowcolor{gray!15}{\ModelName ($w/o\ \bm{R}_k$)} & \underline{56.5} & \textbf{68.5} & \underline{97.8} \\
\bottomrule[1.5pt]
\end{tabular}
\end{table}

Since {\ModelName} is a fundamentally new reasoning architecture, we primarily evaluate its efficacy by training from scratch to ensure that the observed reasoning capabilities stem from the architecture itself rather than external knowledge from pretraining. For a fair comparison, all baselines in the main experiments are trained from scratch using the same data.
More details of experimental settings are provided in Appendix~\ref{append:eva}. We also briefly discuss the pretraining of {\ModelName} in Appendix~\ref{appen:limit}, which we leave for future work. 

\subsubsection{Main Results}\label{sec:main_results}
As shown in Table~\ref{tabel:main} and \ref{tabel:main_3B}, {\ModelName} outperforms existing baselines in accuracy, while providing up to $15.7\times$ speedup compared with traditional transformer-based CoTs. 
Specifically, {\ModelName} (equation) achieves an $8.04$\% accuracy improvement over the best CODI baseline on GSM8K.
%and even beats its 3B version by $1.51$\%.
Beyond in-domain performance, {\ModelName} yields gains of up to $2.30$\% on SVAMP and MultiArith, which demonstrates strong out-of-domain generalization. 
Notably, {\ModelName} also exceeds CoT-SFT in most settings.
%, which provides preliminary evidence that implicit reasoning can reach the level of explicit CoT.
%to the best of our knowledge, 
In particular, when trained with text supervision, {\ModelName} improves accuracy by $4.70$\% on GSM8K. 
These results highlight its ability to execute implicit and deep reasoning effectively. 
%We also analyze performance on harder problems requiring more than 3 reasoning steps in Appendix~\ref{append:hard_gsm8k}, where {\ModelName} shows even larger advantages. 

In terms of efficiency, {\ModelName} not only outperforms token-based CoTs, but also achieves faster inference than many of the 0.5B baselines. 
This underscores that modeling reasoning at the thought level and decoupling the thinking and speaking bring substantial computational advantages. Remarkably, {\ModelName} requires less training time than all 3B models, and even some 0.5B models. These findings demonstrate that our two-stage training scheme is both effective and exceptionally efficient.
% \footnote{The long time of Coconut mainly stems from its original implementation that only supports batch size = 1 in inference.  training time \textcolor{red}{and we find it difficult to modify it to do batch parallelization because xxx.}}. 
%Cosmos scales efficiently and balances accuracy with computational cost.
% \subsection{Ablation Study}\label{section:ablation}
\begin{figure}[t]
  \centering
  \setlength{\abovecaptionskip}{2pt}
  % 调整总宽度和间距，所有图像放置在一行
  \begin{subfigure}[b]{0.55\textwidth}
  \setlength{\abovecaptionskip}{0pt}
  \begin{subfigure}[b]{0.295\textwidth}  % 调整每个子图的宽度
    \centering
    \includegraphics[width=\textwidth]{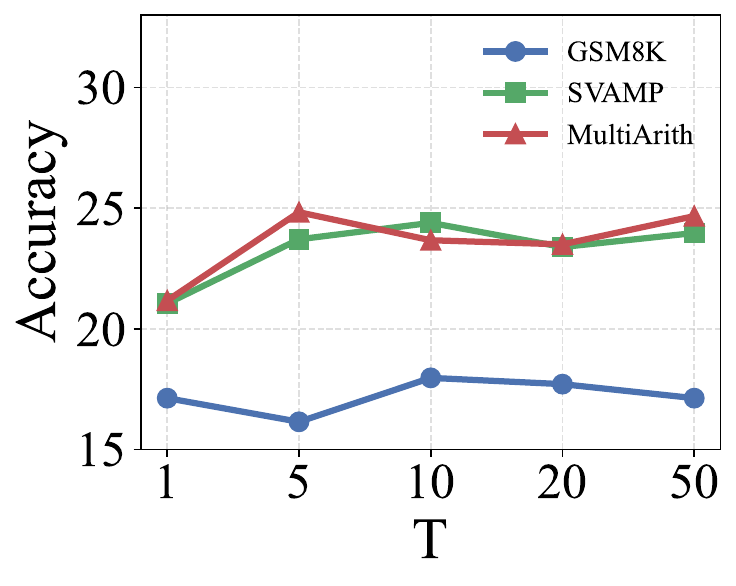}
    %\caption{NL}
  \end{subfigure}
  \hspace{-5pt}
  \begin{subfigure}[b]{0.295\textwidth}  % 调整每个子图的宽度
    \centering
    \includegraphics[width=\textwidth]{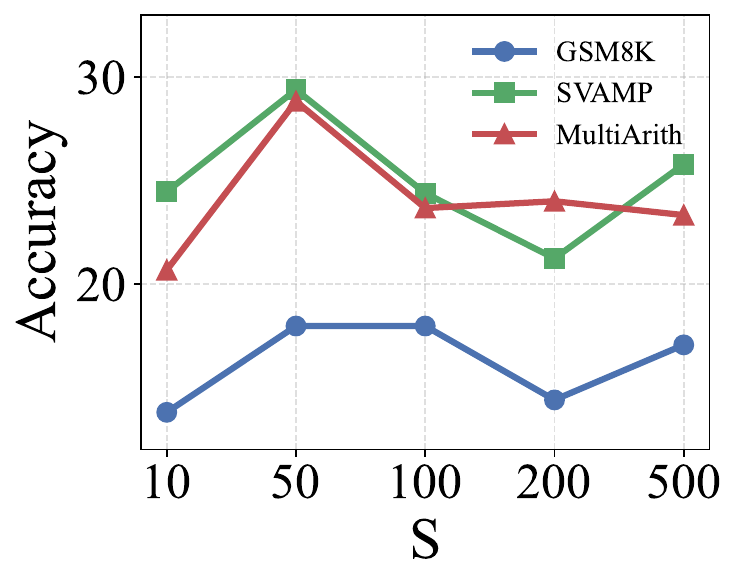}
    %\caption{NL}
  \end{subfigure}
  \hspace{-5pt}
  \begin{subfigure}[b]{0.295\textwidth}  % 调整每个子图的宽度
    \centering
    \includegraphics[width=\textwidth]{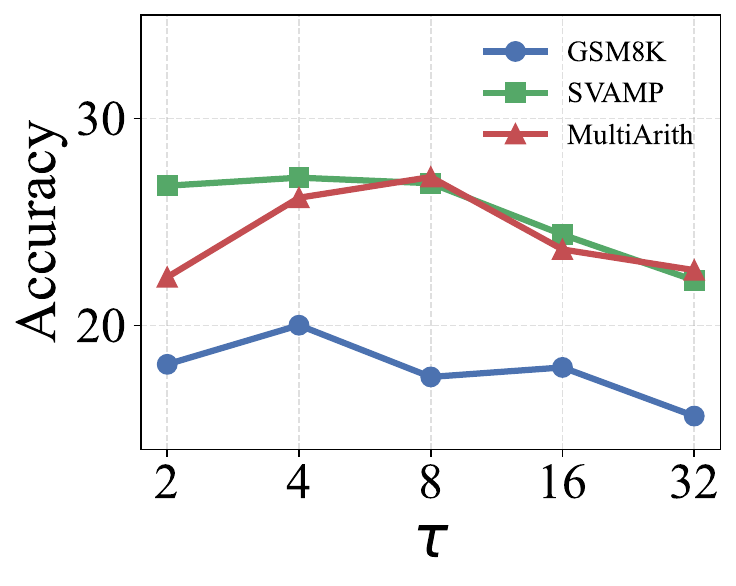}
    %\caption{NL}
  \end{subfigure}
  \caption{{\ModelName} (text)} 
  \end{subfigure}
   \begin{subfigure}[b]{0.55\textwidth}
  \setlength{\abovecaptionskip}{0pt}
  \begin{subfigure}[b]{0.295\textwidth}  % 调整每个子图的宽度
    \centering
    \includegraphics[width=\textwidth]{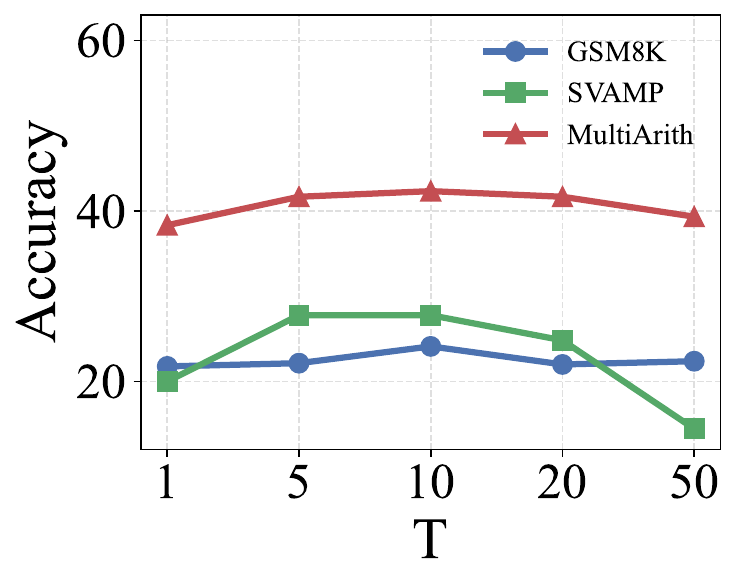}
    %\caption{NL}
  \end{subfigure}
  \hspace{-5pt}
  \begin{subfigure}[b]{0.295\textwidth}  % 调整每个子图的宽度
    \centering
    \includegraphics[width=\textwidth]{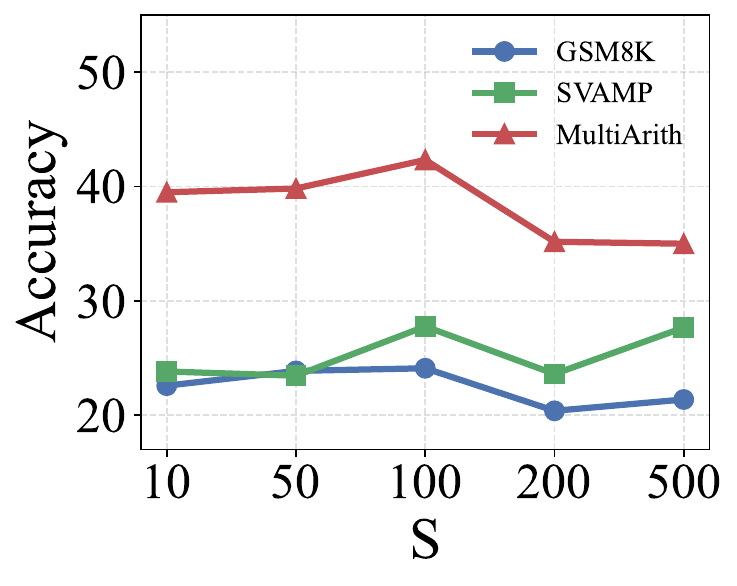}
    %\caption{NL}
  \end{subfigure}
  \hspace{-5pt}
  \begin{subfigure}[b]{0.295\textwidth}  % 调整每个子图的宽度
    \centering
    \includegraphics[width=\textwidth]{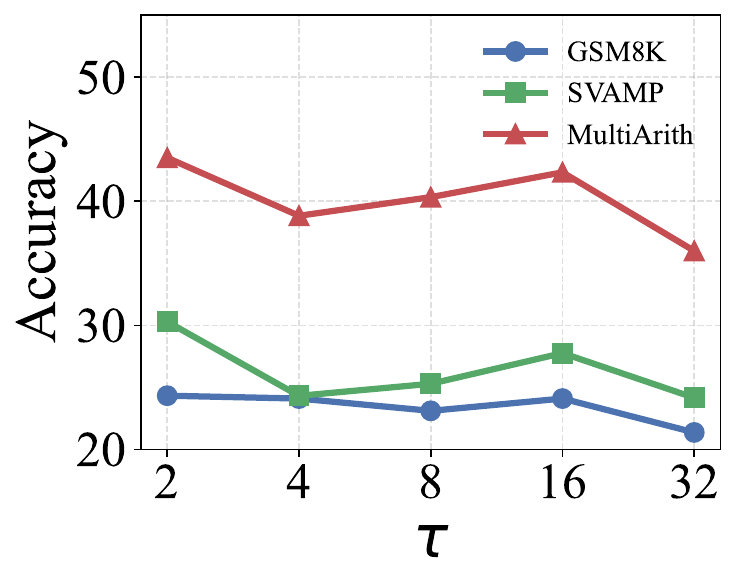}
    %\caption{NL}
  \end{subfigure}
  \caption{{\ModelName} (equation)} 
  \end{subfigure}
  \caption{Performance of {\ModelName} with different numbers of neurons and random variables.}
  \label{fig:delta}
  \vspace{-18pt}
\end{figure}
% \begin{figure*}[t]
%   \centering
%   % 三张小图
% \hspace{10pt} 
%   \begin{subfigure}[b]{0.285\textwidth}
%     \centering
%     \includegraphics[width=1.03\textwidth]{figures/Tr_case.pdf}
%   \end{subfigure}
%   \hspace{5pt}
%   \begin{subfigure}[b]{0.285\textwidth}
%     \centering
%     \includegraphics[width=\textwidth]{figures/Tr_distribution.pdf}
%   \end{subfigure}
%   \hspace{5pt}
%   \begin{subfigure}[b]{0.3\textwidth}
%     \centering
%     \includegraphics[width=0.9\textwidth]{figures/Tr_heatmap1.pdf}
%   \end{subfigure}

%   \caption{\textbf{Left}: $\bm{R}_k$ of step 1 for three cases, \textbf{Middle}: Frequency distribution of all dimensions in $\bm{R}_k$, \textbf{Right}: Correlation heatmap of the top 1\% most frequently activated dimensions.}
%   \label{fig:example_tr}
%   \vspace{-10pt}
% \end{figure*}

\textbf{Ablation Study.} To show the importance of different components of {\ModelName}, we compare it with two variants: $w/o\ \mathbf{Neu}^{deep}$, which removes the deep thinking neurons; $w/o\ \bm{R}_k$, which removes the randomness controller.
%; and $w/o\ sparsity$, which removes the sparsity loss $L^{sparse}_k$.

In Table~\ref{tabel:main}, we observe that removing any of these components leads to a substantial performance drop.
%, and removing $L^{sparse}_k$ even leads to model collapse. 
% This can be attributed to two reasons. 
% On one hand, without the sparsity loss, the random variable $\bm{R}_k$ degenerates into a  dense vector, whose distribution may becomes too complex for the randomness predictor $g$ to learn in the second phase of training. 
% On the other hand, without a proper control, the model may encode excessive semantic information of $y_k$ into $\bm{R}_k$ via the randomness encoder $f$. This information can be easily merged into $\mathbf{Neu}_{k}^{shallow}$ for the reconstruction of $y_k$, creating a training ``shortcut'', which prevents the model from learning genuine reasoning. 
We also find that the relative importance of $\mathbf{Neu}^{deep}$ and $\bm{R}_k$ depends on the supervision type. 
When trained with equations, removing $\bm{R}_k$ has less impact than $\mathbf{Neu}^{deep}$, whereas with text supervision the opposite holds. 
This is because equations carry very little inherent randomness, while texts can exhibit more variability in aspects such as length and format, which will be validated in Section~\ref{section:analysis}. We provide additional ablation studies, including attention direction, initialization strategy, and fixed $\lambda$, in Appendix~\ref{append:ablation}.

\textbf{Attempting Initialization with Pretrained Transformers.}
% In Table~\ref{tab:pretrained_results}, we try to initialize all baselines, and the understanding and speaking layers in {\ModelName} from the Qwen2.5-0.5B checkpoint~\citep{qwen2025qwen25technicalreport}.
% Because the thinking layer of {\ModelName} functions in a very different way from autoregressive transformers, we still train it from scratch.
% Under this setting, {\ModelName} still outperforms all baselines across benchmarks, demonstrating the robustness of our gains beyond training-from-scratch settings. 
% Furthermore, we conduct a more comprehensive experiment where 
In Table~\ref{tab:combined_results}, we try to initialize all layers of {\ModelName} from the pretrained Qwen2.5-0.5B and LLaMA-3.2-1B-Instruct checkpoint, with all baselines initialized accordingly for a fair comparison. However, we find that pretrained transformer weights are not naturally suited for the randomness factor $\bm{R}_k$, as it serves a fundamentally different role from standard tokens. Therefore, we remove $\bm{R}_k$ in this setting to ensure a clean comparison. Under this setting, {\ModelName} continues to achieve superior performance across all three benchmarks. Notably, it achieves substantial gains over CoT on SVAMP (+3.2\% for Qwen; +1.8\% for LLaMA) and MultiArith (+10.0\% for Qwen). These results reflect that our architecture can initialize partial components from existing checkpoints to leverage their pretrained knowledge.
% It is worth noting that baseline performance can vary substantially with different backbones. 
% For instance, the original paper of Coconut reports a 31.7\% accuracy on GSM8K~\citep{hao2024training}. However, when switching to Llama3.2-1B-Instruct, its accuracy drops to 23.1\%~\citep{tan2025think}. We further evaluate all methods with LLaMA-3.2-1B-Instruct as backbone in Appendix~\ref{append:llama}.
% \textcolor{red}{We should probably have an example to show how CoSMos work. It would be the best if we can put it in figure 1? Otherwise can have it in the appendix.}
\subsubsection{Analysis and Discussions}\label{section:analysis}
\textbf{More thinking neurons, better performance?}
We first investigate how the scale of thinking neurons and random variables influences the model's performance by varying the number of deep thinking neurons $T \in \{1,5,10,20,50\}$, shallow thinking neurons $S \in \{10,50,100,200,500\}$, and random variables $\tau \in \{2,4,8,16,32\}$ in Figure~\ref{fig:delta}.
%and Figure~\ref{fig:delta_text} in Appendix, respectively. 
%\textcolor{red}{Have we made it clear that the number of neurons actually have number x hiddensize values?}

For both types of data, as the numbers of neurons $T$ and $S$ increase, the model performance exhibits an initial increase, followed by a period of stabilization or decline.
This suggests that performance is initially constrained by the model’s thinking capacity, but beyond a certain point, adding more neurons may introduce noise or lead to overfitting.

Regarding the scale of randomness $\tau$, we observe different phenomena on text- and equation-based models. For the equation setup, a large $\tau$ generally leads to worse performance, consistent with our earlier hypothesis in Section~\ref{sec:main_results} that equation supervision contains relatively little variability. In contrast, for the text setup, good performance is achieved only with a large $\tau$, indicating that a higher degree of stochasticity is required to adequately capture the diverse variations in natural language.
\begin{figure}[t]
  \centering
  % 三张小图
%\hspace{3pt} 
  \begin{subfigure}[b]{0.23\textwidth}
    \centering
    \includegraphics[width=1\textwidth]{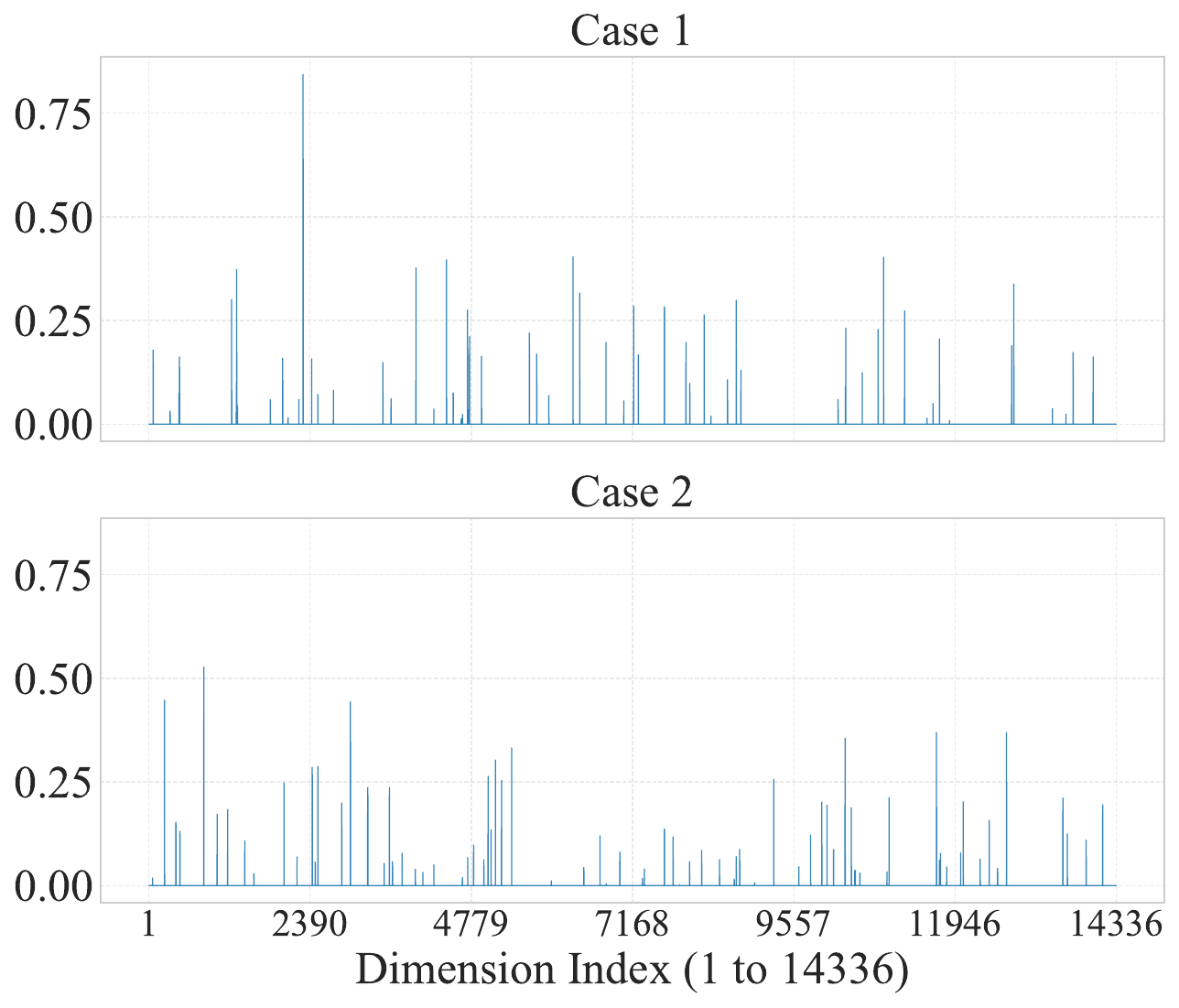}
  \end{subfigure}
  \hspace{-2pt}
  \begin{subfigure}[b]{0.235\textwidth}
    \centering
    \includegraphics[width=\textwidth]{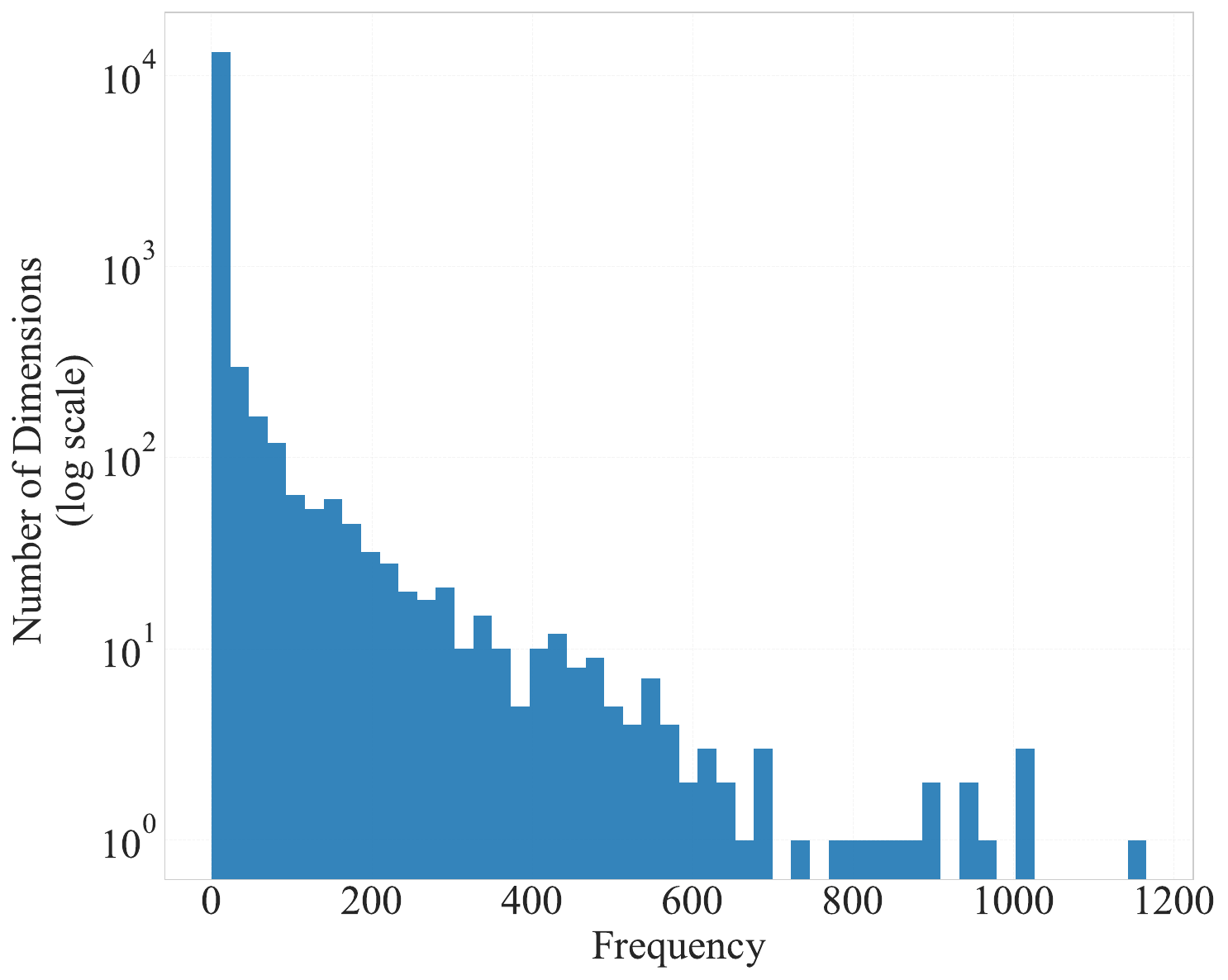}
  \end{subfigure}
  \caption{\textbf{Left}: $\bm{R}_k$ of step 1 for three cases, \textbf{Right}: Frequency distribution of all dimensions in $\bm{R}_k$.}
  \label{fig:example_tr}
  \vspace{-20pt}
\end{figure}
%\textbf{\bm{$Neu^s$} represents the essential part of reasoning.}

\textbf{How does \bm{$\bm{R}_k$} control the randomness of reasoning?} Beyond analyzing random variable $\bm{R}_k$ on synthetic tasks in Section~\ref{sec:syn_results}, we extend our study to real-world reasoning tasks. As shown in Figure~\ref{fig:example_tr}, different data activate different dimensions of $\bm{R}_k$. To quantify this effect, we define dimensions with values greater than $0.1$ as \emph{activated} and count activation frequencies across all samples. The right part of Figure~\ref{fig:example_tr} shows a long-tail pattern: while most dimensions are activated only once, a small subset of dimensions are activated at a very high frequency. This suggests that general randomness information may be controlled by these high-frequency dimensions, whereas case-specific randomness is reflected in the low-frequency ones.
% \begin{figure}[t]
% \centering
% %\setlength{\abovecaptionskip}{2pt}
% \includegraphics[width=0.8\linewidth]{figures/Tr_length_format.pdf}
% \caption{Results of intervening a clique of highly correlated dimensions in $\bm{R}_k$.
% }
% \label{fig:tr_length_format}
% \vspace{-15pt}
% \end{figure}

To verify this, we select the top 1\% most frequently activated dimensions and compute their pairwise Pearson correlation coefficients in Figure~\ref{fig:heat}. Interestingly, we find that several dimensions are highly correlated with each other. In Appendix~\ref{append:vis}, we further visualize this correlation structure as a graph, where we observe a clique of four dimensions that may represent a form of ``combinational control''. Specifically, we identify that these dimensions mainly control two factors: (1) \emph{output length}, i.e., the number of tokens in the reasoning steps, and (2) \emph{output format}, i.e., whether the final answer is preceded by the marker ``$\#\#\#\#$''. To illustrate, we manually intervene on these dimensions by setting their values to $\{0, 0.1, 0.5\}$ for all data samples. Figure~\ref{fig:heat} shows that with higher values, the generated solutions become substantially longer. For example, at $0.5$, the average length increases by $44.62$\%. Meanwhile, the strictness of format diminishes: at $0$ and $0.1$, almost all answers preserve the ``$\#\#\#\#$'' prefix, whereas at $0.5$ the proportion drops sharply to 1.08\%. Crucially, despite these changes, reasoning accuracy remains largely unaffected (red line). These findings indicate that these dimensions independently control randomness factors such as verbosity and format, without encoding the essential reasoning content.

For the case-specific information, we perform a case-by-case analysis in Appendix~\ref{case_dimen}. First, we observe that several dimensions affect the \emph{reasoning depth}. For example, when setting the dimensions $A$ or $C$ to $0.5$, both cases exhibit deeper reasoning at Step~1, completing the original Step 1 and Step 2 simultaneously. Second, we identify distinct functional roles of certain dimensions. In Case 1, dimensions $B$ control whether units are explicitly expressed in the equations (e.g., ``4'' vs. ``4 pens''), whereas in Case 2, dimensions $D$ determine whether numbers are represented as fractions (``$20/100$'') or decimals (``$0.20$''). These findings reflect that our variational module effectively captures nuanced variations in reasoning.
\begin{figure}[t]
  \centering
  % 三张小图
%\hspace{6pt} 
  \begin{subfigure}[b]{0.225\textwidth}
    \centering
    \includegraphics[width=\textwidth]{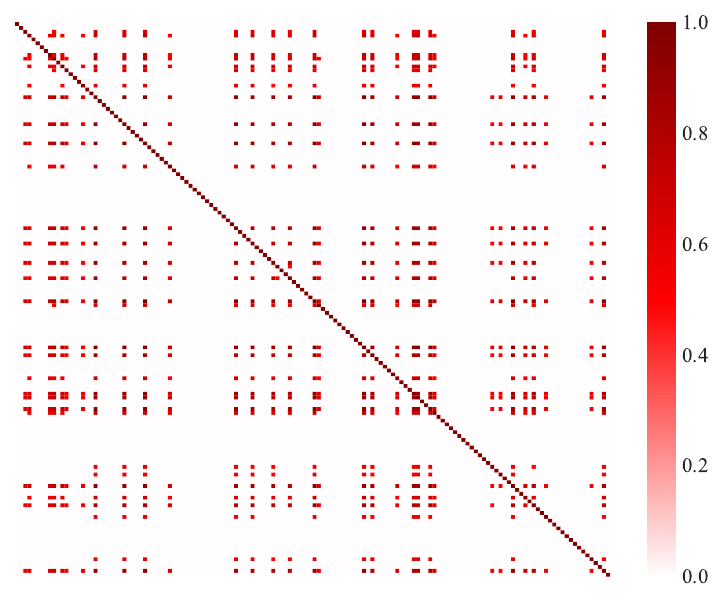}
  \end{subfigure}
  \hspace{-2pt}
  \begin{subfigure}[b]{0.25\textwidth}
    \centering
    \includegraphics[width=1\textwidth]{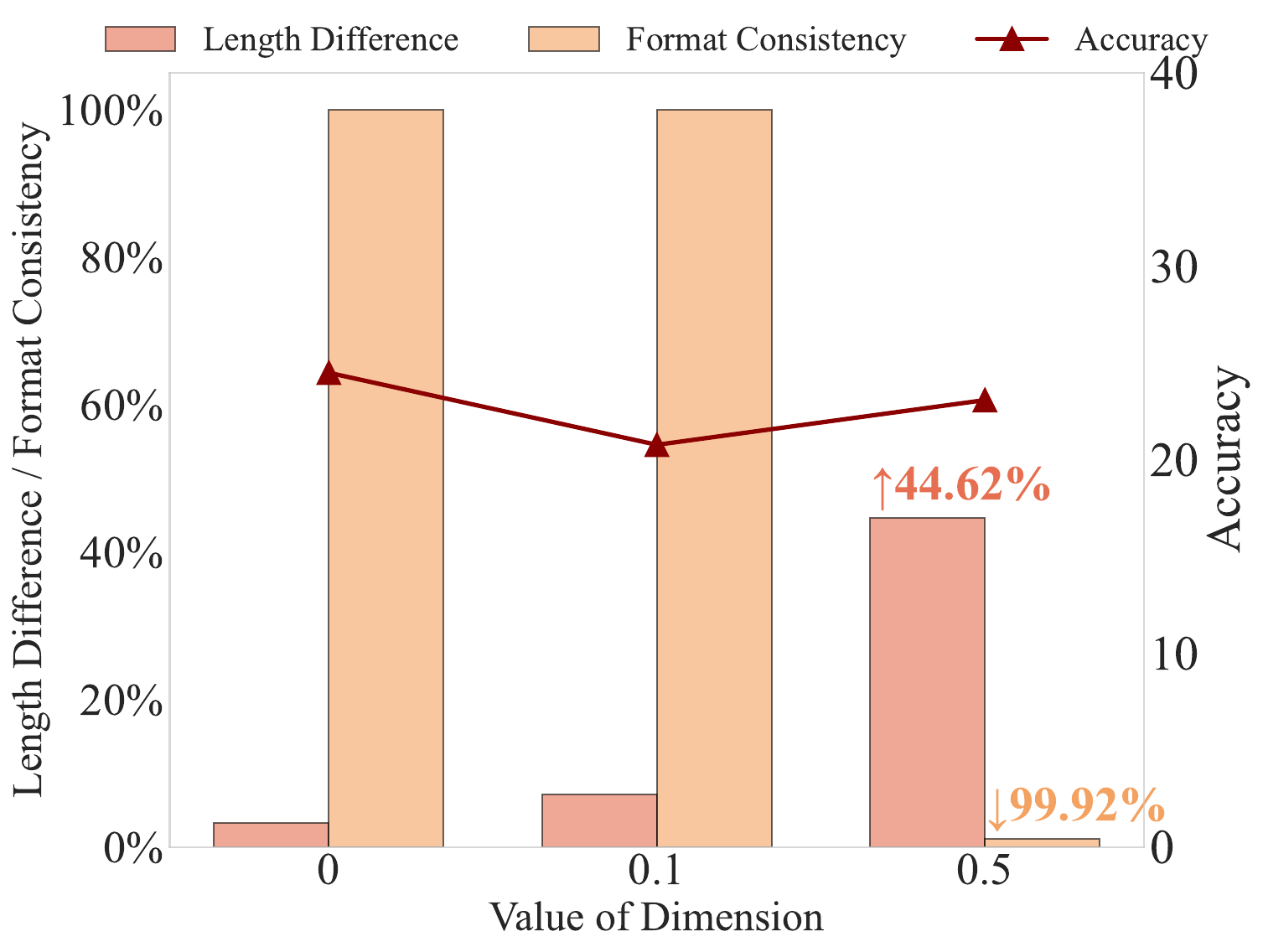}
  \end{subfigure}
  \caption{\textbf{Left}: Correlation heatmap of the top 1\% most frequently activated dimensions in $\bm{R}_k$, \textbf{Right}: Results of intervening a clique of highly correlated dimensions in $\bm{R}_k$.}
  \label{fig:heat}
  \vspace{-15pt}
\end{figure}

\textbf{Potential for non-autoregressive decoding in speaking stage.} A key advantage of {\ModelName} is the explicit separation between thinking and speaking. This means that a clear idea about what to say
%Therefore, we examine whether the speaking stage can support NAR decoding, which has the potential to further improve inference efficiency. Specifically, since our focus is not on optimizing NAR strategies, 
% \begin{wraptable}{r}{0.47\linewidth}
% \caption{Reasoning accuracy of NAR decoding at the speaking stage.}
% \vspace{-5pt}
% \centering
% \small
% \setlength{\tabcolsep}{1.4pt}
% \renewcommand{\arraystretch}{1.1}
% \begin{tabular}{lccc}
% \toprule
% Method & GSM8K & SVAMP & MultiArith \\
% \midrule
% {\ModelName} (text) & 9.02 & 14.96 & 8.33 \\
% {\ModelName} (equation) & 14.25 & 22.82 & 30.12 \\
% \bottomrule
% \end{tabular}
% \label{tab:nar}
% \vspace{-8pt}
% \end{wraptable}
%each reasoning step is constrained to a fixed length of $128$ tokens, which are decoded in parallel without iterative refinement. 
is already formed before entering the speaking stage. As a result, there will be less stochasticity and lower correlation between tokens during sentence decoding, which makes {\ModelName} well-suited for faster non-autoregressive generation. We verify this idea by directly training the speaking layer to generate 128 tokens in one pass. In Table~\ref{tab:nar}, even with such a naive training recipe, {\ModelName} still achieves performance comparable to baselines such as iCoT-SI in Table~\ref{tabel:main}. This finding highlights the potential of {\ModelName} to integrate more advanced NAR strategies, which is an important future direction.

%our model still demonstrates non-trivial reasoning capability. In particular, when trained on the equation data, it still achieves performance comparable to autoregressive baselines such as iCoT-SI. This finding highlights the potential of {\ModelName} to integrate more advanced NAR strategies, which is an important future direction.

\section{Conclusion}
In this work, we introduced {\ModelName}, an improvement of the transformer architecture that performs stepwise reasoning entirely in a continuous space, modeled as a hidden Markov chain. 
Through extensive experiments, we showed that {\ModelName} not only surpasses traditional transformers but also outperforms existing continuous reasoning methods while providing substantial inference speedup. 
%Our analysis further highlights how disentangling thinking from speaking allows for step-level control over stochasticity. 
Looking ahead, our top priority is to verify the effectiveness of {\ModelName} when pretrained on massive corpus and develop dedicated reinforcement learning algorithms for the post-training of {\ModelName}.
%We will also explore more advanced NAR strategies to further improve the efficiency of {\ModelName}. 
For more discussions about this work's limitations and future directions, please refer to Appendix~\ref{appen:limit}.

\nocite{langley00}
\section*{Impact Statement}
This paper presents work whose goal is to advance the field of machine learning. There are many potential societal consequences of our work, none of which we feel must be specifically highlighted here.

\bibliography{example_paper}
\bibliographystyle{icml2026}

%%%%%%%%%%%%%%%%%%%%%%%%%%%%%%%%%%%%%%%%%%%%%%%%%%%%%%%%%%%%%%%%%%%%%%%%%%%%%%%
%%%%%%%%%%%%%%%%%%%%%%%%%%%%%%%%%%%%%%%%%%%%%%%%%%%%%%%%%%%%%%%%%%%%%%%%%%%%%%%
% APPENDIX
%%%%%%%%%%%%%%%%%%%%%%%%%%%%%%%%%%%%%%%%%%%%%%%%%%%%%%%%%%%%%%%%%%%%%%%%%%%%%%%
%%%%%%%%%%%%%%%%%%%%%%%%%%%%%%%%%%%%%%%%%%%%%%%%%%%%%%%%%%%%%%%%%%%%%%%%%%%%%%%
\newpage
\appendix
\onecolumn
\section{Markov Assumption and Long-Range Dependencies}\label{append:longrange}
A natural concern is whether the first-order Markov assumption in {\ModelName} limits its ability to retain information over many reasoning steps. We emphasize that the thought at each step is represented by high-dimensional continuous vectors updated without discretization or information bottleneck, so the model has the inductive capacity to encode all necessary prior information. To empirically validate this, we design a task where the model randomly generates 5 numbers for $K{-}1$ steps and must randomly copy the numbers from one of these prior steps at the $K$-th step, requiring it to retain all $K{-}1$ prior steps' information. As shown in Table~\ref{tab:longrange}, {\ModelName} successfully learns long-range dependencies at $K{=}16$ (88.2\%) and maintains non-trivial performance even at $K{=}32$ and $K{=}64$, where CoT collapses entirely.

\begin{table}[h]
\centering
\small
\caption{Long-range dependency test: accuracy of retaining information from all prior steps.}
\label{tab:longrange}
\begin{tabular}{lccc}
\toprule[1.5pt]
$K$ & 16 & 32 & 64 \\
\midrule[1.2pt]
CoT & 92.5 & 0.0 & 0.1 \\
\cellcolor{gray!15}{\ModelName} & \cellcolor{gray!15}88.2 & \cellcolor{gray!15}8.8 & \cellcolor{gray!15}3.2 \\
\bottomrule[1.5pt]
\end{tabular}
\end{table}

\begin{figure*}[t]
\centering
\setlength{\abovecaptionskip}{5pt}
\includegraphics[width=0.9\linewidth]{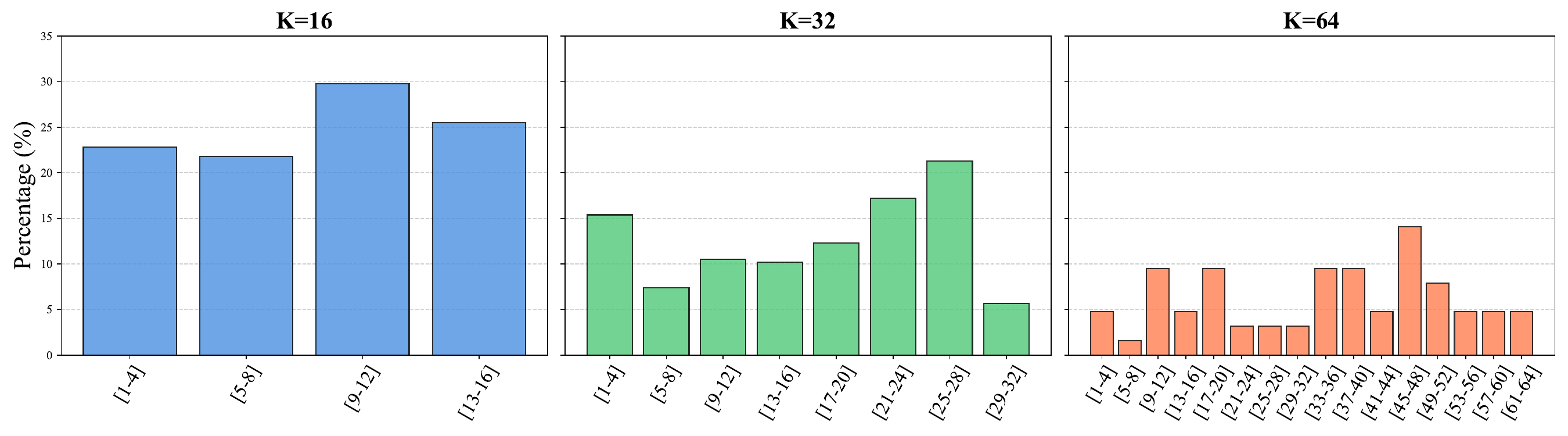}
\caption{Distribution of selected source steps at test time for the long-range dependency task (\%). Each bin covers 4 prior steps.}
\label{figure_distribution_steps}
\vspace{-5pt}
\end{figure*}

To address the concern that the model might be ``cheating'' by always copying from nearby steps, we analyze the distribution of the step selected by the model at test time. The target step to copy is uniformly randomly selected during training, so the model has no incentive to degenerate into a biased selection strategy. Figure~\ref{figure_distribution_steps} shows the test-time selection distribution. At $K{=}16$, the model's selections are well-distributed across all prior step bins. At larger $K$, while the distribution becomes slightly more uneven (as expected for harder tasks), the model still covers all step positions rather than concentrating on a fixed subset, confirming genuine long-range information retention.

\section{Dynamic Weighting Strategy}\label{append:dynamic}
Since it is difficult to predefine a proper balance between the reconstruction loss and the sparsity regularization, we adopt a simple dynamic weighting strategy~\citep{miao2023learning} that directly controls the \emph{expected sparsity level} during training.

Specifically, we introduce a target sparsity $\beta$ and adjust $\lambda$ based on the sparsity $S_{batch}$ on the current training batch $t$:
\begin{equation}
\lambda_{t+1} =
\begin{cases}
1.01 \cdot \lambda_t, & \text{if } S_{batch} > \beta, \\
0.99 \cdot \lambda_t, & \text{if } S_{batch} \le \beta .
\end{cases}
\end{equation}
We initialize $\lambda_0$ to $10^{-4}$ to avoid overly strong regularization in the early training stage. Empirically, we find that this strategy stably drives the sparsity level to converge to the target value $\beta$, while avoiding overly aggressive sparsity optimization that could otherwise cause the reconstruction loss to prematurely fall into poor local minima.
\begin{table*}[t]
\caption{
Results of models \textbf{trained from scratch}. All continuous reasoning baselines (3B) are trained on equation data, following their original papers. We report the best continuous models in bold and the runner-ups with underline.
}
\centering
\small
\setlength{\tabcolsep}{7pt}
\renewcommand{\arraystretch}{1.}
\begin{tabular}{l|l|ll|ll|ll}
\toprule[1.5pt]
 & & \multicolumn{2}{c|}{GSM8K} & \multicolumn{2}{c|}{SVAMP} & \multicolumn{2}{c}{MultiArith} \\
\cmidrule(lr){3-4} \cmidrule(lr){5-6} \cmidrule(lr){7-8}
Model & Train (h) & Acc (\%) & Test (s)  & Acc (\%) & Test (s)  & Acc (\%) & Test (s) \\
\midrule[1.2pt]
\multicolumn{8}{c}{\textit{\textbf{Training Data: Text}}} \\
\midrule
CoT-SFT & ${1.81 \scriptstyle \pm 0.00}$ & ${12.66 \scriptstyle \pm 0.20}$ & ${177.37 \scriptstyle \pm 1.91}$ & ${24.24 \scriptstyle \pm 0.50}$ & ${537.78 \scriptstyle \pm 12.14}$ & ${23.44 \scriptstyle \pm 1.40}$ & ${58.20 \scriptstyle \pm 2.50}$ \\
\cellcolor{gray!15}{\ModelName} & \cellcolor{gray!15}${1.13 \scriptstyle \pm 0.01}$ & \cellcolor{gray!15}${\textbf{17.97} \scriptstyle \pm 0.96}$ & \cellcolor{gray!15}${17.53 \scriptstyle \pm 0.29}$ & \cellcolor{gray!15}${\textbf{24.39} \scriptstyle \pm 1.10}$ & \cellcolor{gray!15}${34.14 \scriptstyle \pm 0.49}$ & \cellcolor{gray!15}${\textbf{23.67} \scriptstyle \pm 0.53}$ & \cellcolor{gray!15}${5.90 \scriptstyle \pm 0.17}$ \\
\midrule[1.2pt]
\multicolumn{8}{c}{\textit{\textbf{Training Data: Equation}}} \\
\midrule
CoT-SFT &${1.23 \scriptstyle \pm 0.00}$ & ${22.70 \scriptstyle \pm 0.36}$ & ${54.07 \scriptstyle \pm 2.16}$ & ${27.25 \scriptstyle \pm 0.46}$ & ${125.47 \scriptstyle \pm 7.67}$ & ${50.20 \scriptstyle \pm 1.60}$& ${10.73 \scriptstyle \pm 0.17}$ \\
iCoT-SI&${2.40 \scriptstyle \pm 0.10}$ & ${14.08 \scriptstyle \pm 0.67}$ & ${13.23 \scriptstyle \pm 0.06}$ & ${16.26 \scriptstyle \pm 0.27}$ & ${29.02 \scriptstyle \pm 0.44}$ & ${19.67 \scriptstyle \pm 0.58}$ & ${3.49 \scriptstyle \pm 0.05}$ \\
Coconut &${12.62 \scriptstyle \pm 1.06}$ & ${9.95 \scriptstyle \pm 0.04}$ & ${127.35 \scriptstyle \pm 0.44}$ & ${13.53 \scriptstyle \pm 0.11}$& ${735.14 \scriptstyle \pm 3.80}$ & ${8.78 \scriptstyle \pm 0.19}$ & ${26.56 \scriptstyle \pm 0.66}$ \\
CoLaR & ${2.21 \scriptstyle \pm 0.00}$ & ${\underline{22.60} \scriptstyle \pm 0.19}$ & ${38.13 \scriptstyle \pm 0.21}$ & ${\underline{25.83} \scriptstyle \pm 0.22}$ & ${129.73 \scriptstyle \pm 0.96}$ & ${\underline{41.83} \scriptstyle \pm 0.44}$ & ${18.93 \scriptstyle \pm 0.73}$ \\
% CODI & ${3.48 \scriptstyle \pm 0.00}$ &${1.24 \scriptstyle \pm 0.36}$ & ${12.91 \scriptstyle \pm 0.14}$ & ${1.12 \scriptstyle \pm 0.17}$ & ${34.18 \scriptstyle \pm 0.24}$ & ${1.20 \scriptstyle \pm 0.68}$& ${4.94 \scriptstyle \pm 0.14}$ \\
% CODI$^{\dagger}$ & -- & 18.40 & -- & 26.09 & -- & 41.41 & -- \\
\cellcolor{gray!15}{\ModelName} & \cellcolor{gray!15}${1.03 \scriptstyle \pm 0.01}$ & \cellcolor{gray!15}${\textbf{24.11} \scriptstyle \pm 0.97}$ & \cellcolor{gray!15}${17.76 \scriptstyle \pm 0.04}$ & \cellcolor{gray!15}${\textbf{27.77} \scriptstyle \pm 0.32}$ & \cellcolor{gray!15}${34.02 \scriptstyle \pm 0.11}$ & \cellcolor{gray!15}${\textbf{42.33} \scriptstyle \pm 0.56}$ & \cellcolor{gray!15}${6.02 \scriptstyle \pm 0.28}$ \\
\bottomrule[1.5pt]
\end{tabular}
\label{tabel:main_3B}
%\vspace{-8pt}
\end{table*}
% \section{Overview of Synthetic Tasks}\label{append:synthetic}
% In Figure~\ref{task_all}, we present three examples for the synthetic tasks.
\section{Details of Implementation and Evaluation}\label{append:eva}
In {\ModelName}, the understanding layer, speaking layer, and randomness encoder are realized by stacking standard transformer layers, using the same model configuration as Qwen2.5-0.5B. 
The thinking layer, in contrast, is a specialized module that updates a set of neurons and random variables. 
Therefore, it is a stack of transformer layers with bi-directional attention to allow each neuron to take the states of all other neurons into consideration. % While basic computations such as linear projections and layer normalization can still follow transformer conventions, but with attention mechanism should be modified to be bi-directional to support the information flow in thought transition. 
In other words, we do not modify the internal computations of transformers, nor is it the goal of our work. Instead, we repurpose the standard transformer to achieve fundamentally different roles in {\ModelName}. In particular, through the design of the thinking layer, we demonstrate that a transformer-like structure can perform one-step reasoning within a single forward pass.

The numbers of deep and shallow neurons are set to $T=10$, $S=100$. The scaling factor of randomness variable $\bm{R}_k$ is set to $\tau=16$. {\ModelName} is trained by AdamW with learning rate $1\mathrm{e}{-4}$ and weight decay $0.01$. The training batch size is 256, and both training phases run for 10 epochs. To simplify prototyping, the number of thinking steps is fixed to $K=3$, although our model also supports dynamically deciding this number. For text data, each period (``.'') is treated as a separate step. If a sample contains more than three steps, the original steps are evenly grouped and recombined into three steps. For fair comparison, the CoT-SFT baseline is trained for 20 epochs. All experiments are conducted on a server with 8 H200 GPUs.

In our evaluation, Training Time is defined as the wall-clock time required to complete one epoch of training. To ensure a fair comparison, for baselines that require pre-training an explicit CoT model, the reported training time is the sum of (i) the time to train the explicit CoT model for one epoch, and (ii) the time to train their own model for one epoch. For our model, the reported training time is the sum of the two training phases, each measured over one epoch. Test Time is the total time required to infer all test samples on a single GPU, using a batch size of 64. However, the original implementation of Coconut only supports inference with batch size = 1. We attempted to modify it to 64, but observed no significant improvement in efficiency. Therefore, for Coconut we report the time of its default setting, i.e., 4 GPUs with batch size = 1.

\section{Additional Ablation Studies}\label{append:ablation}
We provide additional results to examine the sensitivity of {\ModelName} to architectural and hyperparameter choices. (1) \textbf{Attention direction.} We compare bidirectional attention (our default) with unidirectional attention in the thinking layer. As shown in Table~\ref{tab:ablation_combined}, bidirectional attention consistently outperforms unidirectional attention, confirming that allowing neurons to attend to all other neurons is beneficial for reasoning. (2) \textbf{Initialization strategy.} The performance differences between our default initialization with Xavier and Normal initialization for the thinking layer are marginal across all three benchmarks, indicating that {\ModelName} is robust to initialization choice. (3) \textbf{Fixed $\lambda$.} Although {\ModelName} adopts an adaptive dynamic weighting strategy for $\lambda$, we also evaluate fixed values for completeness. From Table~\ref{tab:ablation_combined}, fixing $\lambda$ to any of the tested values degrades performance compared to the adaptive strategy, motivating the need for dynamic balancing.

\begin{table}[h]
\centering
\small
\caption{Ablation on attention direction and initialization strategy (equation data).}
\label{tab:ablation_combined}
\begin{tabular}{llccc}
\toprule[1.5pt]
Ablation & Setting & GSM8K & SVAMP & MultiArith \\
\midrule[1.2pt]
Attention & Bidirectional (default) & 24.11 & 27.77 & 42.33 \\
Attention & Unidirectional & 22.74 & 25.35 & 38.82 \\
\midrule[1.2pt]
$\lambda$ & $10^{-4}$ & 21.32 & 23.98 & 40.54 \\
$\lambda$ & $5{\times}10^{-4}$ & 20.83 & 23.90 & 39.43 \\
$\lambda$ & $10^{-3}$ & 20.83 & 23.82 & 36.39 \\
\bottomrule[1.5pt]
\end{tabular}
\end{table}
\section{Visualization of Dimension Correlation Structure}\label{append:vis}
\begin{figure}[!t]
\centering
\includegraphics[width=0.83\linewidth]{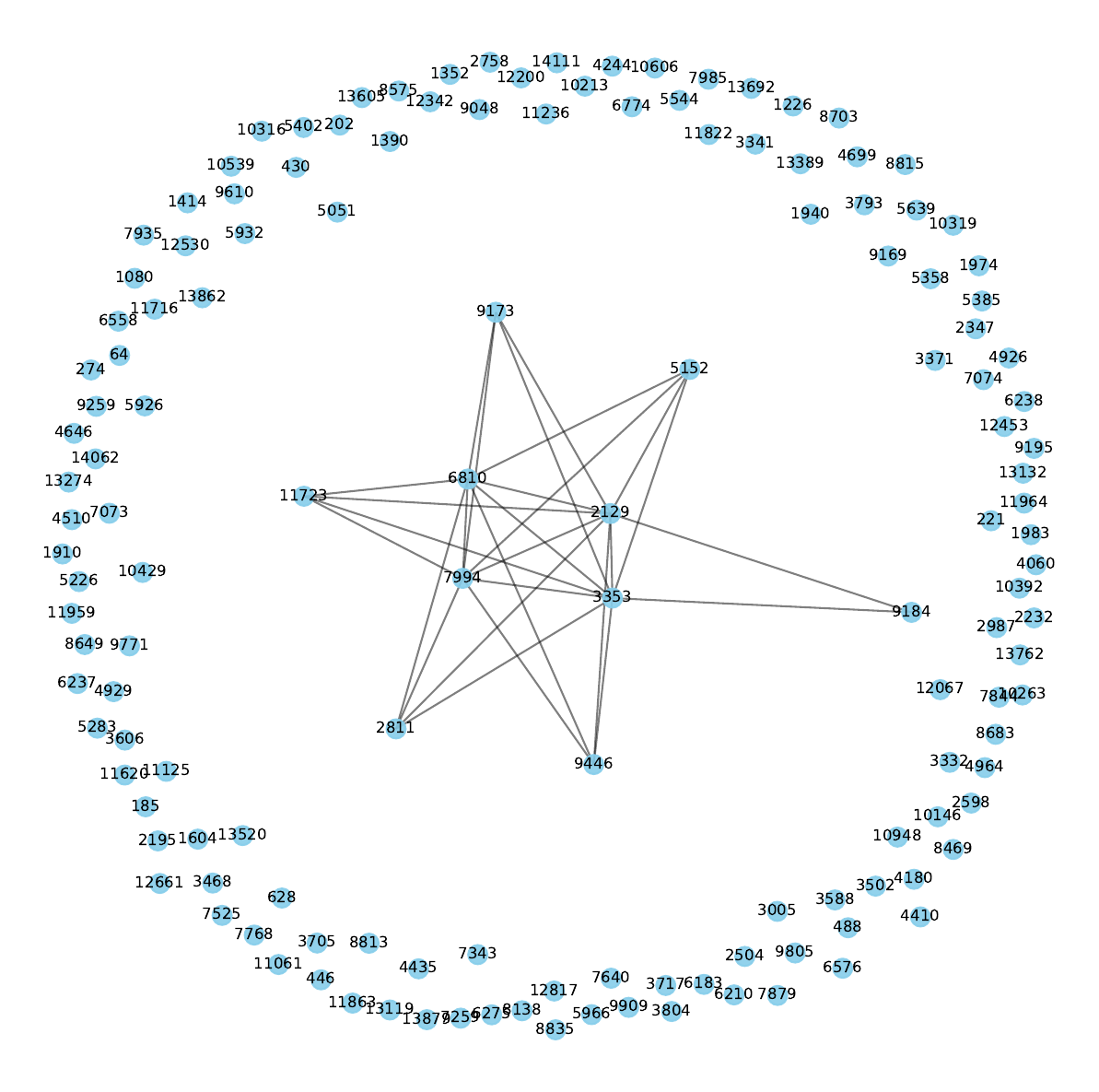}
\caption{Correlation graph of the top 1\% most frequently activated dimensions in $\bm{R}_k$.}
\label{figure_graph}
\vspace{-10pt}
\end{figure}

Figure~\ref{figure_graph} visualizes the correlation graph described in Section~\ref{section:analysis}. In this graph, each node represents one of the top $1\%$ most frequently activated dimensions, and an edge is drawn between two nodes if their Pearson correlation exceeds $0.9$. From the figure, we can clearly observe a clustered structure, suggesting that some dimensions tend to be co-activated in a highly consistent manner. Notably, within the cluster we further identify a clique formed by dimensions \{2129, 3353, 6810, 7994\}, which indicates that these dimensions jointly control aspects of general randomness. 
% \begin{figure}[t]
%   \centering
%   \begin{subfigure}[b]{\textwidth}
%     \centering
%     % 三张小图
%     \begin{subfigure}[b]{0.3\textwidth}
%       \centering
%       \includegraphics[width=\textwidth]{figures/TT-eq-line.pdf}
%     \end{subfigure}
%     \hspace{5pt}
%     \begin{subfigure}[b]{0.3\textwidth}
%       \centering
%       \includegraphics[width=\textwidth]{figures/TS-eq-line.pdf}
%     \end{subfigure}
%     \hspace{5pt}
%     \begin{subfigure}[b]{0.3\textwidth}
%       \centering
%       \includegraphics[width=\textwidth]{figures/TR-eq-line.pdf}
%     \end{subfigure}
%   \end{subfigure}

%   \caption{Performance of {\ModelName} (equation) with different numbers of neurons and random variables.}
%   \label{fig:delta_text}
% \end{figure}
\begin{figure*}[t]
\centering
\setlength{\abovecaptionskip}{5pt}
\includegraphics[width=0.9\linewidth]{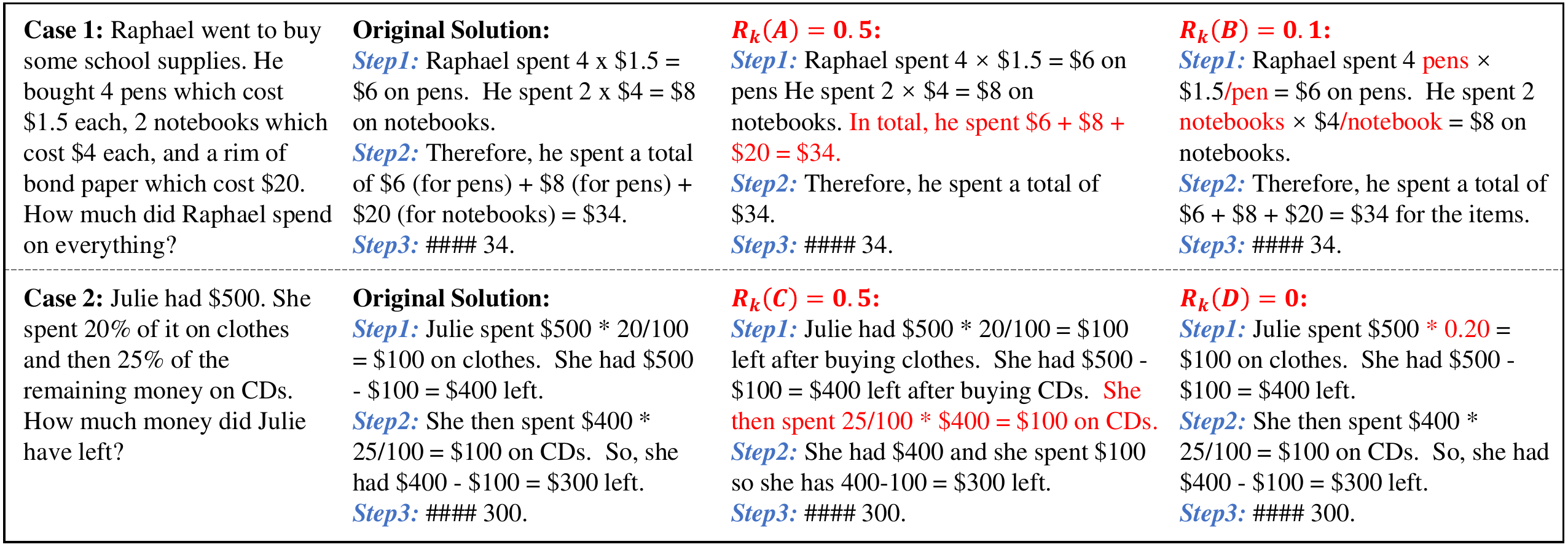}
\caption{Case study of controlling four sets of dimensions in $\bm{R}_k$ ($k=1,...,K$). For each case, the difference of model output after intervention is highlighted in red.}
\label{figure_r_case}
\vspace{-5pt}
\end{figure*}
\section{Case Study}\label{case_dimen}
Figure~\ref{figure_r_case} shows how dimensions in $\bm{R}_k$ control specific information. We highlight in red the differences between the controlled outputs and the original solutions. The dimension sets are: $A$=\{ 199,  3505,  5283,  7879,  9259,  9468,  9485, 11725\}, $B$=\{4060\}, $C$=\{1379,  1583,  5475,  7017,  7074,  7521,   8683,  9635, 10460, 10539, 11716, 12012, 13337\}, $D$=\{1226\}.

As illustrated in Section~\ref{section:analysis}, certain dimensions affect the \emph{reasoning depth}, such as $A$ and $C$. The difference between these two dimensions may arise because dimension $A$ primarily affects the final equation, whereas dimension $C$ influences the penultimate equation. 
This indicates that the stored information in random variables goes beyond surface-level linguistic style, and can in fact control diverse reasoning pathways. 
Other dimensions control finer-grained aspects of the solutions, such as the explicit expression of units ($B$) or numerical format ($D$).

\begin{table}[!t]
\centering
\small
\caption{Accuracy of NAR decoding at the speaking stage.}
\begin{tabular}{lccc}
\toprule
Method & GSM8K & SVAMP & MultiArith \\
\midrule
{\ModelName} (text) & 9.02 & 14.96 & 8.33 \\
{\ModelName} (equation) & 14.25 & 22.82 & 30.12 \\
\bottomrule
\end{tabular}
\label{tab:nar}
%\vspace{-15pt}
\end{table}
\section{Limitations and Future Directions}\label{appen:limit}
The following are some limitations in this work, which also point to promising directions for future research. 
First, the experiments in this paper primarily focus on mathematical reasoning tasks. This choice is motivated by the fact that transformer-based reasoning has proven to be most effective in this domain~\citep{spraguecot}, and most existing reasoning models are also benchmarked under mathematical scenarios. 
Second, our evaluation is conducted on relatively basic mathematical problems, rather than more challenging datasets such as MATH or AIME. 
On one hand, our goal in this paper is to demonstrate the effectiveness and potential of {\ModelName} as a continuous reasoning structure, rather than pursuing leaderboard-oriented results. On the other hand, tackling complex mathematical problems may require additional techniques such as reinforcement learning, which we leave for future work. 
Third, as discussed in Section~\ref{section:exp_setup}, latent thinking represents a new paradigm for transformers. To fully realize its potential, we need to explore corresponding pretraining methods and data, enabling the model to acquire broader and more robust capabilities across diverse domains.

Beyond these limitations, our study also opens several avenues for future work. First, and most importantly, is to develop a pre-training methodology for {\ModelName}. The key challenge may be the definition and segmentation of reasoning steps, as such large, unstructured corpora lack explicit, step-by-step annotations. One potential solution is to treat each sentence as a single reasoning step. Under this setup, the pre-training objective can be formulated as given the preceding sentences (analogous to the question in our paper), the model needs to generate the subsequent $K$ sentences. It is compatible with our existing training scheme, which could be scaled up. Second, as validated in Section~\ref{section:analysis}, our random variable enables step-level control over stochasticity, offering a new perspective for exploration in reinforcement learning. Unlike conventional token-level sampling, our approach allows sampling reasoning pathways at a higher level, potentially making exploration more efficient. Third, our framework shows that randomness can be simulated ``once'' rather than being iteratively modeled, which may inspire new directions for diffusion language models. Last but not least, our work demonstrates the feasibility of disentangling thinking from speaking, suggests that speaking can be treated as a pure translation process. Therefore, it is worth trying to introduce more efficient decoding techniques, especially NAR decoding, to further accelerate inference.
\section{Statement on the Use of LLMs}
We used LLMs to improve the clarity, grammar, and fluency of our paper. The LLMs were not involved in any technical idealizations, derivations, or generation of research content. We have carefully reviewed all content generated by LLMs to ensure that no factual or technical inaccuracies were introduced, and the scientific integrity of the work is fully maintained.
%%%%%%%%%%%%%%%%%%%%%%%%%%%%%%%%%%%%%%%%%%%%%%%%%%%%%%%%%%%%%%%%%%%%%%%%%%%%%%%
%%%%%%%%%%%%%%%%%%%%%%%%%%%%%%%%%%%%%%%%%%%%%%%%%%%%%%%%%%%%%%%%%%%%%%%%%%%%%%%

\end{document}